\documentclass[11pt]{article}

\usepackage{acl}

\usepackage{times}
\usepackage{latexsym}
\usepackage[T1]{fontenc}
\usepackage[utf8]{inputenc}
\usepackage{microtype}
\usepackage{inconsolata}
\usepackage{graphicx}
\usepackage{booktabs}
\usepackage{amsmath}
\usepackage{amssymb}
\usepackage{multirow}
\usepackage{xcolor}
\usepackage{url}
\usepackage{hyperref}
\usepackage{subcaption}
\usepackage{dblfloatfix}
\usepackage{enumitem}
\usepackage{orcidlink}

\title{\textbf{On-Policy Distillation Meets Off-Policy GRPO:}\\
\mdseries Training Compact Instruction-Following Rerankers}

\author{
Vignesh Prabhakar \orcidlink{0000-0002-5459-8015}\quad
Jialing Pan \orcidlink{0009-0005-5240-3068}\quad
Anil Babu Ankisettipalli \orcidlink{0009-0008-1702-4259} \\
SAP Labs, Palo Alto, CA, USA \\
\texttt{\{vignesh.prabhakar01,joyce.pan01,anil.babu.ankisettipalli\}@sap.com}
}

\begin{document}
\maketitle

\begin{abstract}
Compact instruction-following rerankers are attractive for deployment,
but conventional distillation pipelines typically train students by
offline imitation of teacher outputs on a fixed set of examples,
constraining supervision to the teacher's observed ranking space.
We revisit reranker distillation through the lens of reinforcement
learning.

We propose a two-stage framework combining off-policy teacher
optimization with on-policy student distillation. In Stage~1, a 4B
teacher reranker is strengthened with off-policy GRPO using LLM-judge
feedback on 88K instruction-following examples. In Stage~2, a compact
1B student samples rankings from its own policy and receives soft
teacher-derived rewards on those rankings, coupling student exploration
with knowledge transfer.

Our strongest gains appear under distribution shift. On MAIR-11, the
original 11-subset, 869-query evaluation, the proposed student reaches
0.7670 nDCG@6, outperforming offline listwise KD by +4.6 points.
Controlled comparisons against offline pairwise RankNet KD and
on-policy GKD show that neither changing the offline distillation
objective nor moving teacher-distribution matching on-policy reproduces
the performance of reward-based on-policy distillation over
student-sampled rankings. The advantage persists on MAIR-Full:
across all 126 tasks and 9,356 queries, the proposed method obtains
the highest task-macro point estimates among the evaluated distillation
variants, reaching 0.6808 nDCG@6 and 0.7865 MRR@6. It also exceeds
two released 7B RL-trained rerankers on the comparable MAIR-11
evaluation, while the same Stage~2 training procedure consistently
improves three architecturally distinct alternative student backbones.
On the 9,861-query validation benchmark, the resulting 1B reranker
achieves 0.7624 nDCG@6 while providing a favorable
quality--efficiency tradeoff relative to larger alternatives.
\end{abstract}

\section{Introduction}

Neural rerankers are now a core component of modern retrieval systems,
providing fine-grained relevance estimation beyond first-stage retrieval
pipelines~\citep{nogueira2019passage,khattab2020colbert,lin2021pretrained}.
Recent work extends reranking to \emph{instruction-following} settings,
where users specify ranking criteria in natural language rather than
through implicit query intent alone~\citep{asai2023task,weller2024followir,oh2024instructir}.
This capability is especially important in enterprise and assistant-facing
retrieval, where ranking decisions may depend on temporal constraints,
source preferences, audience requirements, or other multi-factor
instructions.

Despite this progress, training \emph{compact} instruction-following
rerankers that generalize robustly remains difficult. Large proprietary
rerankers offer strong performance but are costly and latency-sensitive
in production. More fundamentally, existing student training approaches
typically reduce distillation to offline imitation: the student is
trained to match teacher-provided labels or score distributions on a
fixed set of examples, constraining the student to the teacher's
observed ranking space and potentially transferring teacher-specific
biases under distribution shift.

We revisit reranker distillation from a reinforcement learning
perspective. Our central hypothesis is that effective capability
transfer requires \emph{student-driven exploration}: the student should
generate rankings from its own policy and learn from teacher-derived
rewards on those rankings. We propose a two-stage pipeline:
\textbf{Stage~1} trains a 4B teacher using off-policy GRPO with
LLM-judge feedback; \textbf{Stage~2} distills into a compact 1B
reranker using on-policy GRPO with soft teacher rewards.

On-policy distillation has recently emerged as a strong post-training
paradigm for reasoning and code generation, where the student receives
teacher supervision on its own generated token
trajectories~\citep{agarwal2024onpolicy,lu2025onpolicydistillation,qwen3}.
We carry this paradigm into a different action space:
\emph{rankings} rather than token sequences. A ranking is a permutation
over candidate documents, and teacher supervision is a soft reward over
sampled rankings rather than a logit distribution over next tokens.

Our contributions:

\begin{itemize}

\item \textbf{Reward-based on-policy distillation for ranking.}
    We formulate reranker distillation as policy learning over ranking
    permutations: the student samples rankings from its own
    Plackett--Luce policy and receives soft teacher rewards on those
    student-generated rankings.

\item \textbf{Controlled mechanism evidence.}
    On MAIR-11, the proposed method improves over offline listwise KD
    by +4.6 nDCG@6 points. Additional controls using offline pairwise
    RankNet KD (A10) and on-policy GKD (A9) show that neither pairwise
    offline transfer nor on-policy teacher-distribution matching
    reproduces the performance of reward-based on-policy distillation.

\item \textbf{Broad OOD and architectural generalization.}
    The proposed method obtains the highest task-macro point estimates
    among evaluated distillation variants on all 126 MAIR tasks, while
    the same Stage~2 procedure consistently improves three
    architecturally distinct alternative student backbones. The 1B
    student also exceeds two released 7B RL-trained rerankers on the
    comparable MAIR-11 evaluation.

\item \textbf{Compact deployment.}
    The resulting 1B reranker reaches 0.7624 nDCG@6 on the
    9,861-query validation benchmark while substantially reducing
    inference latency relative to 4B alternatives.

\end{itemize}

\section{Related Work}

\paragraph{Instruction Following Reranking.}
Neural rerankers improve retrieval quality by rescoring candidate
documents, often using cross-encoder
architectures~\citep{nogueira2019passage,lin2021pretrained}. Recent work
extends this to instruction-following settings where models condition on
natural-language ranking
criteria~\citep{asai2023task,weller2024followir,oh2024instructir},
supported by benchmarks such as \textsc{FollowIR} and
\textsc{InstructIR}. Systems such as ZeRank-2~\citep{zerank2model}
demonstrate that instruction-following reranking benefits from RL-based
alignment, while compact open rerankers such as
Llama-Nemotron-Rerank-1B-v2~\citep{nemotronrerank1bv2} provide strong
efficiency-oriented starting points. We build on this setting but ask
how a compact reranker should acquire instruction-following behavior
from a stronger teacher without inheriting the teacher's fixed
prediction distribution.

\paragraph{Prompt-based and LLM-based Reranking.}
Promptriever~\citep{weller2024promptriever} makes dense retrieval
models instruction-sensitive through prompting. RankGPT and related
methods treat ranking as an inference-time prompting
problem~\citep{qin2024rankgpt,ma2023zeroshotlistwise}, while
RankVicuna and RankZephyr show that open-source LLMs can be effective
zero-shot listwise rerankers~\citep{pradeep2023rankvicuna,pradeep2023rankzephyr}.
We differ in goal: rather than large prompted LLMs at inference time,
we train a compact 1B reranker through a two-stage RL pipeline.

\paragraph{Reinforcement Learning for Ranking.}
RL has become standard for optimizing non-differentiable objectives in
language generation and alignment~\citep{ouyang2022training,rafailov2023direct}.
GRPO~\citep{shao2024deepseekmath} provides efficient policy-gradient
optimization via group-normalized advantages. Prior work in offline RL
emphasizes that policies trained on fixed data can be limited by
distributional coverage, whereas on-policy interaction can correct
errors on states induced by the current policy~\citep{levine2020offline}.
In reranking, RL has mainly been used as an alignment stage for large
rerankers or in off-policy settings. Our method uses GRPO in two
distinct roles: off-policy for the teacher and on-policy for student
distillation.

\paragraph{Knowledge Distillation for Retrieval.}
Knowledge distillation transfers behavior from large teachers to smaller
students~\citep{hinton2015distilling}. In retrieval, prior work commonly
performs offline distillation from fixed teacher
scores~\citep{hofstatter2021efficiently}. Born-again distillation has
shown that students can sometimes outperform teachers via a
regularization effect from soft
supervision~\citep{furlanello2018born}. LLM-as-judge
methods~\citep{zheng2023judging,dubois2024alpacaeval} provide scalable
soft reward signals richer than binary labels. We depart from standard
offline KD by treating distillation as policy learning: the student
samples rankings from its own policy and receives teacher-derived
rewards, enabling exploration that offline KD does not.

\paragraph{Positioning.}
Our method lies at the intersection of two previously separate
directions. Generalized on-policy knowledge distillation trains
students on self-generated trajectories using teacher-distribution
matching~\citep{agarwal2024onpolicy}, while recent RL rerankers such
as Rank-R1, REARANK, and ERank optimize ranking behavior directly
using reward signals~\citep{zhuang2026rankr1,zhang2025rearank,
cai2026erank}. Prior reranker-distillation approaches instead
predominantly transfer teacher information offline or address
training--inference alignment through alternative objectives
~\citep{choi2024rradistill,huang2025gumbel,xu2025distillation}.
Our method occupies the intersection left open by these directions:
\emph{reward-based on-policy distillation over ranking permutations}.
The student samples complete rankings from its own Plackett--Luce
policy, and a fixed teacher assigns a scalar soft reward to each
student-sampled permutation. A9 makes this distinction empirical by
retaining student-generated rankings while replacing permutation-level
reward supervision with GKD-style teacher-distribution matching.
\section{Method}

\paragraph{Model naming conventions.}
\textbf{ZeRank-2} denotes the 4B teacher
backbone~\citep{zerank2model}. \textbf{Teacher GRPO} refers to
ZeRank-2 after Stage~1 off-policy RL. \textbf{Base-1B} denotes the
Llama-Nemotron-Rerank-1B-v2 student
backbone~\citep{nemotronrerank1bv2}, and \textbf{Distilled-1B} denotes
the final student after Stage~2 on-policy distillation.

\subsection{Problem Setup}

\begin{figure*}[t]
\centering
\includegraphics[width=0.93\textwidth]{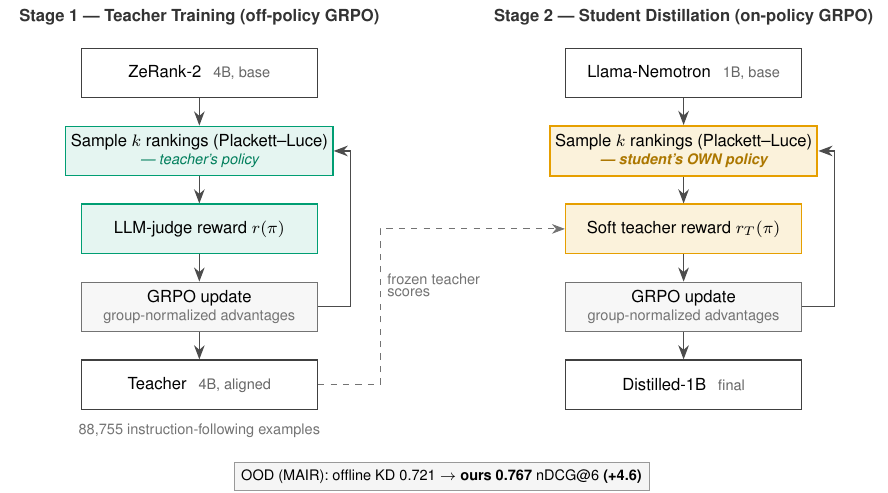}
\caption{
Two-stage training pipeline. Both stages use the same GRPO machinery over Plackett–Luce–sampled rankings; they differ only in whose policy generates the rankings and what provides the reward. Stage 1 aligns a 4B teacher with LLM-judge feedback (off-policy with respect to the eventual student). Stage 2 distills into a 1B student on-policy: the student samples rankings from its own policy and receives soft teacher-derived rewards, which drives the +4.6 nDCG@6 OOD improvement over offline KD (§5.2).}
\label{fig:architecture}
\end{figure*}

Given a query $q$, instruction $i$, and candidate document set
$D=\{d_1,\dots,d_n\}$, the reranker defines a scoring function over
documents and induces a distribution over rankings. Our training
pipeline has two stages (Figure~\ref{fig:architecture}). Stage~1 trains
a stronger teacher using off-policy GRPO with LLM-judge feedback.
Stage~2 distills the teacher into a compact student using on-policy
GRPO, where the student samples rankings from its own policy and
receives teacher-derived soft rewards. The central difference from
offline distillation is that supervision is generated on the student's
sampled rankings rather than on a fixed teacher-generated distribution.

\subsection{Stage 1: Off-Policy GRPO Teacher Training}

For each training example $(q,i,D)$, the teacher induces a distribution
over rankings and we sample candidate rankings
$\pi^{(1)},\dots,\pi^{(k)}$ using Plackett--Luce sampling. An LLM judge
assigns a reward $r(\pi^{(j)})$ to each sampled ranking. Following
GRPO~\citep{shao2024deepseekmath}, we compute group-normalized
advantages
\begin{equation}
A(\pi^{(j)})=\frac{r(\pi^{(j)})-\mu}{\sigma},
\end{equation}
where $\mu,\sigma$ are the mean and standard deviation of the sampled
rewards. The teacher is updated with
\begin{align}
\mathcal{L}_{\text{teacher}} ={}&
-\mathbb{E}_{\pi\sim p_\theta}\!\left[A(\pi)\log p_\theta(\pi)\right]
\nonumber\\
&+\beta_{\text{KL}}\mathrm{KL}(p_\theta\|p_{\text{ref}})
-\alpha_{\text{ent}}H(p_\theta).
\end{align}

\subsection{Stage 2: On-Policy GRPO Distillation}

In Stage~2, the student samples rankings from its \emph{own} policy:
$\pi^{(1)},\dots,\pi^{(k)}\sim p_{\theta_S}$. Let $s_j$ denote the
student score for candidate $d_j$. These scores directly parameterize
a Plackett--Luce distribution,

\begin{equation}
p_{\theta_S}(\pi)
=
\prod_{m=1}^{n}
\frac{\exp(s_{\pi_m})}
{\sum_{j\notin\pi_{<m}}\exp(s_j)}.
\end{equation}

For each candidate $d_j$, the fixed Stage~1 teacher produces a scalar
utility
\begin{equation}
t_j = z^{T}_{\mathrm{Yes}}(q,i,d_j)/5.0,
\end{equation}
where $z^{T}_{\mathrm{Yes}}$ is the teacher's final-position
\texttt{Yes}-token logit. We rank-normalize the teacher utilities to
obtain soft relevance values and score each student-sampled
permutation using nDCG@6 under those teacher-derived relevances:

\begin{equation}
r_T(\pi)=\mathrm{nDCG@6}(\pi;\mathrm{rel}_T).
\end{equation}

Thus, the teacher does not generate the student's training rankings;
it \emph{evaluates} rankings sampled by the student. We compute the
group-normalized teacher advantage

\begin{equation}
A_T(\pi^{(j)})
=
\frac{r_T(\pi^{(j)})-\mu_T}{\sigma_T},
\end{equation}

where $\mu_T$ and $\sigma_T$ are the mean and standard deviation of
teacher rewards within the sampled group.

For the auxiliary regularizers, we use the first-selection
Plackett--Luce marginals

\begin{equation}
q_S=\mathrm{softmax}(s),
\qquad
q_T=\mathrm{softmax}(t),
\end{equation}

and optimize

\begin{align}
\mathcal{L}_{\text{student}} ={}&
-\mathbb{E}_{\pi\sim p_{\theta_S}}
 \left[A_T(\pi)\log p_{\theta_S}(\pi)\right]
\nonumber\\
&+\lambda_{\text{KL}}\mathrm{KL}(q_S\|q_T)
-\alpha_{\text{ent}}H(q_S).
\end{align}

The KL term is therefore computed over the first-selection
Plackett--Luce marginal rather than over the combinatorial distribution of complete permutations. The entropy bonus is analogously computed from the student's first-selection marginal. The key property of the method is that supervision is evaluated on the student's own sampled ranking space, making distillation an on-policy learning process.

\subsection{Hypothesis: Why On-Policy Distillation Generalizes}

Offline distillation supervises the student only on a fixed teacher
distribution, encouraging imitation of rankings the teacher already
prefers. On-policy distillation instead evaluates teacher-derived
supervision on rankings sampled from the student's own policy, expanding
the effective coverage of the training signal. We therefore expect the
gap between offline KD and on-policy distillation to be smallest
in-distribution and larger under distribution shift---exactly the
pattern observed in our experiments.

\subsection{Roles of Objective Components}

Our ablations show that the dominant learning signal comes from
policy-gradient optimization with teacher rewards. The KL and entropy
terms are optional stabilizers with limited effect on in-distribution
performance but can improve OOD robustness. We retain them in the
final model and analyze their effect in
Section~\ref{sec:ablation}.

\section{Experimental Setup}

\subsection{Datasets}

Our instruction-following training and validation benchmark combines
eight datasets spanning web search, code, mathematics, news, and
multi-hop retrieval. The training split contains 88,755 examples and
the held-out validation benchmark contains 9,861 queries
(Table~\ref{tab:data}).

\begin{table}[t]
\centering
\small
\begin{tabular}{lrr}
\toprule
\textbf{Dataset} & \textbf{Train} & \textbf{Val} \\
\midrule
FollowIR (TREC) & 445 & 49 \\
InstructIR & 8,915 & 991 \\
InfoSearch & 4,296 & 477 \\
InfIR (MS MARCO) & 34,883 & 3,876 \\
InfIR (MetaMath) & 6,394 & 710 \\
InfIR (LeetCode) & 2,286 & 254 \\
InfIR (Robust04) & 1,765 & 196 \\
M-BEIR (WebQA) & 29,771 & 3,308 \\
\midrule
\textbf{Total} & \textbf{88,755} & \textbf{9,861} \\
\bottomrule
\end{tabular}
\caption{Instruction-following reranking data used for training and
validation.}
\label{tab:data}
\end{table}

For out-of-distribution evaluation, we use
MAIR~\citep{sun2024mair}, a heterogeneous instructed-retrieval
benchmark containing 126 tasks across six domains. We report two
complementary evaluation settings.

\textbf{MAIR-11} denotes the original 11-subset, 869-query evaluation
used for our controlled ablations and external-model comparisons.
It spans ad hoc retrieval (Core\_2017, DD\_2016), FAQ matching
(Quora), scientific and biomedical evidence retrieval (SciFact,
SciDocs, Trec-Covid, NFCorpus, LitSearch), financial retrieval
(FiQA), and argumentative retrieval (ArguAna, Touche). These subsets
were selected to reflect heterogeneous retrieval settings relevant to
our target deployment scenario rather than to optimize performance on
any single domain.

\textbf{MAIR-Full} extends the evaluation to all 126 MAIR tasks,
comprising 9,356 queries in total. Because MAIR tasks vary
substantially in size, we report task-macro averages for this
full-benchmark analysis.

\paragraph{Candidate pools and truncation.}
We use benchmark-provided candidate pools where available and construct BM25 candidate pools otherwise. Candidate pools are created once during preprocessing, frozen, and reused identically across all systems and ablations, preventing first-stage retrieval variation from confounding reranker comparisons. Model inputs are truncated to a maximum length of 512 tokens. In the training and validation splits, typical candidate pools contain 3--18 documents per query (10th--90th percentile), with mean pool size 7.5 and median 3. The full observed range is 2--21 documents in training and 2--20 in validation.

\subsection{Models and Baselines}

Our teacher model is \textbf{ZeRank-2}~\citep{zerank2model} (4B).
Our primary student backbone is
\textbf{Llama-Nemotron-Rerank-1B-v2}~\citep{nemotronrerank1bv2}.

We compare against: (i) internal references (Base-1B, ZeRank-2 base,
Teacher GRPO); (ii) controlled training ablations including supervised
BCE, offline listwise KD (A1), off-policy GRPO distillation (A2),
on-policy GRPO distillation (A3), distillation from the unaligned base
teacher (A8), on-policy GKD (A9), RankNet pairwise KD (A10),
objective-component removals, and hyperparameter sweeps; and
(iii) external baselines including BGE Reranker v2 M3,
Jina Reranker v2, Qwen3-Reranker-4B, Cohere Rerank v3.5,
Cohere Rerank v4.0-fast,
Rank-R1-7B~\citep{zhuang2026rankr1}, and
REARANK-7B~\citep{zhang2025rearank}. We exclude Jina, BGE, and
Cohere from MAIR evaluation due to potential train--test contamination.
Rank-R1-7B and REARANK-7B are evaluated zero-shot, without any
fine-tuning or adaptation on MAIR, using the same frozen MAIR-11
candidate pools as our controlled comparisons. This isolates
reranking quality while holding the first-stage candidate set fixed
across methods.

A9 performs on-policy GKD-style teacher-distribution matching on
student-sampled rankings, whereas A10 performs offline pairwise
RankNet distillation from teacher preferences. Together with A1 and
A3, these controls disentangle the sampling policy from the form of
teacher supervision.

\subsection{Training Configuration}

\paragraph{Stage~1.}
Off-policy GRPO with LLM-judge rewards: group size $k=8$, learning rate
$5\times10^{-6}$ with 3\% warmup, batch size 64, 10 epochs.

\paragraph{Stage~2.}
On-policy GRPO with teacher-derived soft rewards: group size $k=8$,
temperature 1.0, KL weight 1.0, entropy coefficient 0.01, learning rate
$2\times10^{-6}$ with 3\% warmup, batch size 64, 3 epochs. Set~B
varies these hyperparameters. All training on NVIDIA H200 GPUs.

\paragraph{LLM Judge Validation.}
Because Stage~1 relies on judge-derived rewards, we validate the
signal in two ways. Five strong LLM judges show high inter-judge
agreement on relevance ordering and on answer containment, and the
Stage~1 judge reaches 92.6\% agreement with human relevance
annotations (174/188 examples, Cohen's $\kappa=0.84$), with most
disagreements being false negatives. Full figures and per-dataset
breakdowns are provided in Appendix~\ref{sec:appendix_judge}.

\subsection{Evaluation Metrics}

We report \textbf{nDCG@6} and \textbf{MRR@6} as primary ranking
metrics~\citep{jarvelin2002cumulated,chapelle2009expected}.

For the validation benchmark, aggregate metrics are query-micro
averages. We compute 95\% confidence intervals using 10,000
percentile-bootstrap resamples of the pooled per-query metric values,
sampling queries with replacement without stratification.

For MAIR-11, aggregate metrics are query-micro averages over all 869
queries, computed identically to the validation benchmark; confidence
intervals likewise use 10,000 percentile-bootstrap resamples of the
pooled per-query metric values. For MAIR-Full, whose 126 tasks vary
substantially in size, we instead report task-macro averages: we first
compute the mean metric within each task and then average equally
across tasks, with confidence intervals obtained from 10,000
percentile-bootstrap resamples of the resulting per-task means,
sampling tasks with replacement without stratification.

\section{Main Results}

\subsection{Instruction-Following Reranking Performance}

Table~\ref{tab:main_results} reports aggregate performance on the
9,861-query validation benchmark. \textbf{Distilled-1B} achieves the
best overall nDCG@6 among all evaluated models while remaining compact.

\begin{table*}[t]
\centering
\small
\setlength{\tabcolsep}{9pt}
\begin{tabular}{lccccc}
\toprule
\textbf{Model} & \textbf{Params} &
\textbf{nDCG@6} & \textbf{SD} &
\textbf{MRR@6} & \textbf{SD} \\
\midrule
Qwen3-Reranker-4B
& 4B & 0.6564 [0.649, 0.664] & 0.3821
& 0.6387 [0.631, 0.647] & 0.4055 \\

Base-1B (no training)
& 1B & 0.6972 [0.690, 0.705] & 0.3657
& 0.6639 [0.656, 0.672] & 0.3909 \\

ZeRank-2 base (4B)
& 4B & 0.7222 [0.715, 0.730] & 0.3516
& 0.7056 [0.698, 0.714] & 0.3704 \\

BGE Reranker v2 M3
& 568M & 0.7310 [0.724, 0.739] & 0.3697
& 0.7057 [0.698, 0.714] & 0.3897 \\

Teacher GRPO (4B)
& 4B & 0.7422 [0.735, 0.750] & 0.3716
& 0.7256 [0.718, 0.734] & 0.3904 \\

Cohere Rerank v3.5
& -- & 0.7459 [0.7387, 0.7533] & 0.3711
& 0.7265 [0.7190, 0.7343] & 0.3889 \\

REARANK-7B
& 7B & 0.7486 [0.7411, 0.7538] & 0.3626
& 0.7338 [0.7242, 0.7376] & 0.3766 \\

Cohere Rerank v4.0-fast
& -- & 0.7494 [0.7422, 0.7569] & 0.3708
& 0.7325 [0.7251, 0.7404] & 0.3875 \\

Rank-R1-7B
& 7B & 0.7506 [0.7432, 0.7552] & 0.3722
& 0.7351 [0.7266, 0.7394] & 0.3827 \\

Jina Reranker v2
& 278M & 0.7605 [0.753, 0.768] & 0.3737
& 0.7497 [0.742, 0.758] & 0.3884 \\

\textbf{Distilled-1B (ours)}
& \textbf{1B} & \textbf{0.7624 [0.755, 0.770]} & 0.3730
& 0.7475 [0.740, 0.755] & 0.3885 \\
\bottomrule
\end{tabular}
\caption{Main validation results on 9,861 instruction-following
queries. Confidence intervals are bootstrap 95\% CIs with
10,000 resamples.}
\label{tab:main_results}
\end{table*}

Three patterns stand out. First, RL-based training substantially
improves over the untrained 1B backbone, raising nDCG@6 from 0.6972
to 0.7624. Second, the distilled student exceeds both the untrained
4B teacher backbone and the Stage~1 teacher. We treat the +4.6-point
gain over offline KD on MAIR-11, rather than the raw teacher--student
gap, as the cleaner measure of the distillation contribution because
the Stage~1 teacher exhibits concentrated OOD failures discussed in
Section~\ref{sec:ood}. Third, Distilled-1B remains competitive with
strong external rerankers despite its compact size: it exceeds Cohere
v3.5, Cohere v4.0-fast, Rank-R1-7B, and REARANK-7B in validation
nDCG@6, while Jina v2 obtains a slightly higher MRR@6 point estimate.

\subsection{Out-of-Distribution Generalization on MAIR}
\label{sec:ood}

\begin{table*}[t]
\centering
\small
\setlength{\tabcolsep}{10pt}
\begin{tabular}{lcccc}
\toprule
\textbf{Model} & \textbf{nDCG@6} & \textbf{SD} &
\textbf{MRR@6} & \textbf{SD} \\
\midrule
Base-1B (no training)
& 0.7119 [0.698, 0.726] & 0.2150
& 0.7771 [0.762, 0.792] & 0.2250 \\
Teacher GRPO (4B)
& 0.6880 [0.674, 0.702] & 0.2150
& 0.7312 [0.716, 0.747] & 0.2350 \\
REARANK-7B
& 0.7315 [0.717, 0.746] & 0.2180
& 0.7987 [0.783, 0.814] & 0.2300 \\
Rank-R1-7B
& 0.7342 [0.720, 0.748] & 0.2120
& 0.8092 [0.794, 0.824] & 0.2220 \\
A1: Offline KD
& 0.7212 [0.707, 0.735] & 0.2080
& 0.7860 [0.771, 0.801] & 0.2200 \\
A2: Off-policy GRPO
& 0.7356 [0.722, 0.749] & 0.2050
& 0.8035 [0.789, 0.818] & 0.2150 \\
A8: On-policy GRPO from base ZeRank-2
& 0.7412 [0.728, 0.754] & 0.1980
& 0.8036 [0.790, 0.817] & 0.2050 \\
\textbf{A3: On-policy GRPO from Teacher (ours)}
& 0.7670 [0.755, 0.779] & 0.1850
& 0.8289 [0.816, 0.842] & 0.1900 \\
A4: No KL term
& 0.7688 [0.757, 0.781] & 0.1820
& 0.8282 [0.816, 0.841] & 0.1880 \\
A5: No entropy
& 0.7665 [0.754, 0.779] & 0.1860
& 0.8301 [0.817, 0.843] & 0.1900 \\
A6: Hard labels
& 0.7365 [0.723, 0.750] & 0.2080
& 0.8000 [0.786, 0.814] & 0.2150 \\
A9: On-policy GKD
& 0.7386 [0.726, 0.751] & 0.1920
& 0.8088 [0.796, 0.822] & 0.1950 \\
A10: RankNet pairwise KD
& 0.7416 [0.729, 0.754] & 0.1900
& 0.8128 [0.800, 0.826] & 0.1920 \\
\bottomrule
\end{tabular}
\caption{OOD results on MAIR-11, the original 11-subset,
869-query evaluation. Detailed ablation results are provided in
Appendix~\ref{sec:appendix_mair}.}
\label{tab:mair_compact}
\end{table*}

One feature of Table~\ref{tab:mair_compact} deserves direct comment.
Teacher GRPO reaches 0.7422 nDCG@6 on the validation benchmark but
only 0.6880 on MAIR-11, falling below the untrained Base-1B (0.7119).
This degradation is concentrated rather than uniform: Teacher GRPO
outperforms Base-1B on 7 of the 11 MAIR-11 subsets, but large failures
on LitSearch and Touche dominate its aggregate decline. The distilled
student (A3) therefore does not uniformly outperform its teacher;
a substantial part of its aggregate improvement comes from repairing
these concentrated OOD failures.

On MAIR-11, A3 substantially outperforms off-policy sampling (A2),
offline listwise KD (A1), on-policy GKD (A9), RankNet pairwise KD
(A10), and distillation from the unaligned base teacher (A8).
In particular, neither on-policy teacher-distribution matching (A9)
nor offline pairwise RankNet distillation (A10) reproduces the OOD
performance of reward-based on-policy distillation. The A3--A1 gap
is especially notable because the two variants are nearly tied on the
validation benchmark, whereas A3 improves over offline KD by
+4.6 nDCG@6 points on MAIR-11.

A3 also exceeds both released 7B RL-trained rerankers on this
comparable evaluation, reaching 0.7670 nDCG@6 and 0.8289 MRR@6
versus 0.7342/0.8092 for Rank-R1-7B and 0.7315/0.7987 for
REARANK-7B.

We additionally evaluate the applicable distillation variants on
MAIR-Full, comprising all 126 tasks and 9,356 queries. A3 achieves
the highest task-macro point estimates for both metrics, reaching
0.6808 nDCG@6 and 0.7865 MRR@6, compared with 0.6657/0.7565
for A9, 0.6564/0.7512 for A1, and 0.6521/0.7502 for A10.
Complete results are reported in
Appendix~\ref{sec:appendix_mair_full126}.

\section{Ablation Study}
\label{sec:ablation}

\begin{figure*}[t]
    \centering
    \begin{subfigure}[t]{0.465\textwidth}
        \centering
        \includegraphics[width=\textwidth]{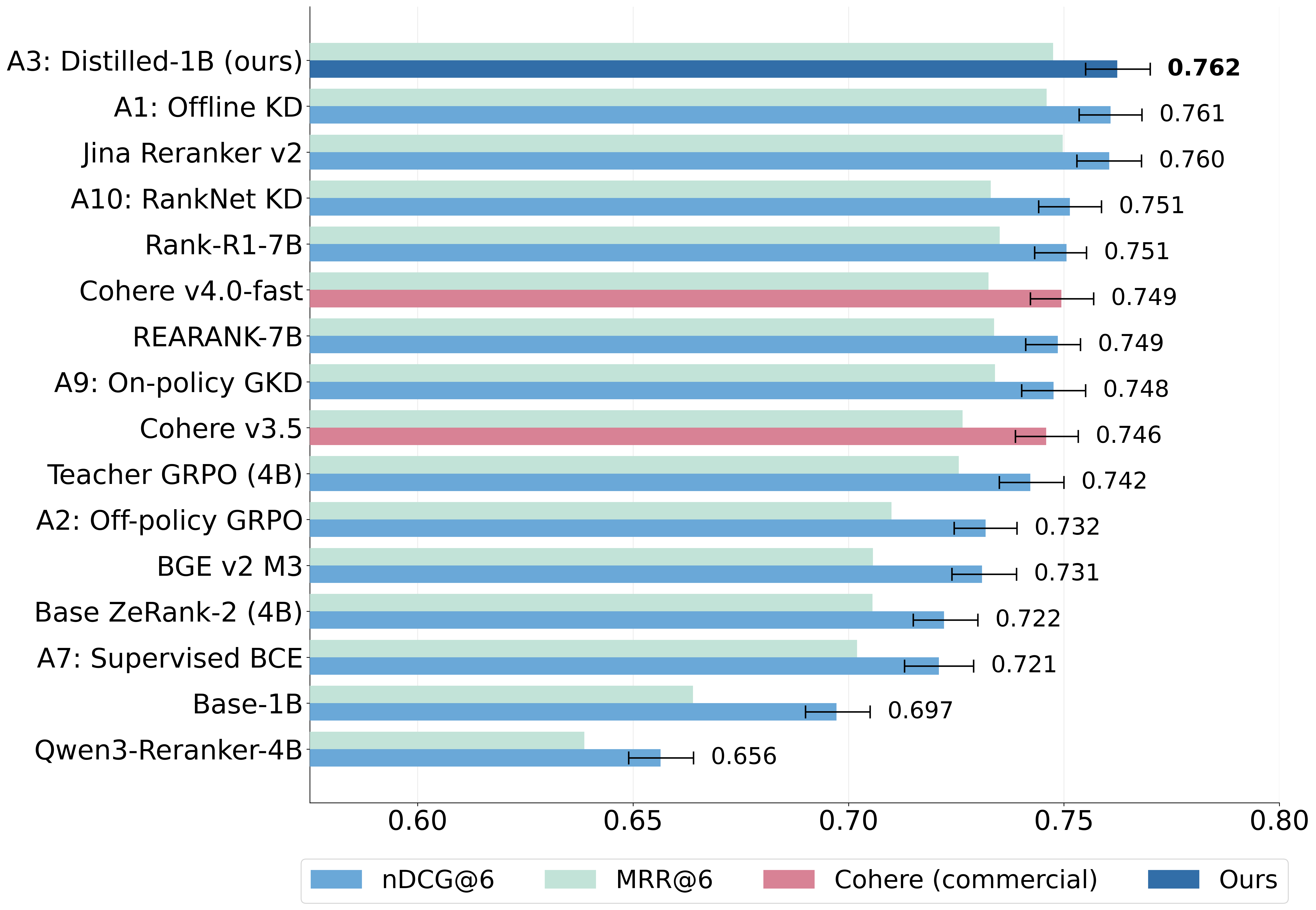}
        \caption{Validation comparison across major baselines and ablations.}
        \label{fig:setA_bar}
    \end{subfigure}
    \hfill
    \begin{subfigure}[t]{0.465\textwidth}
        \centering
        \includegraphics[width=\textwidth]{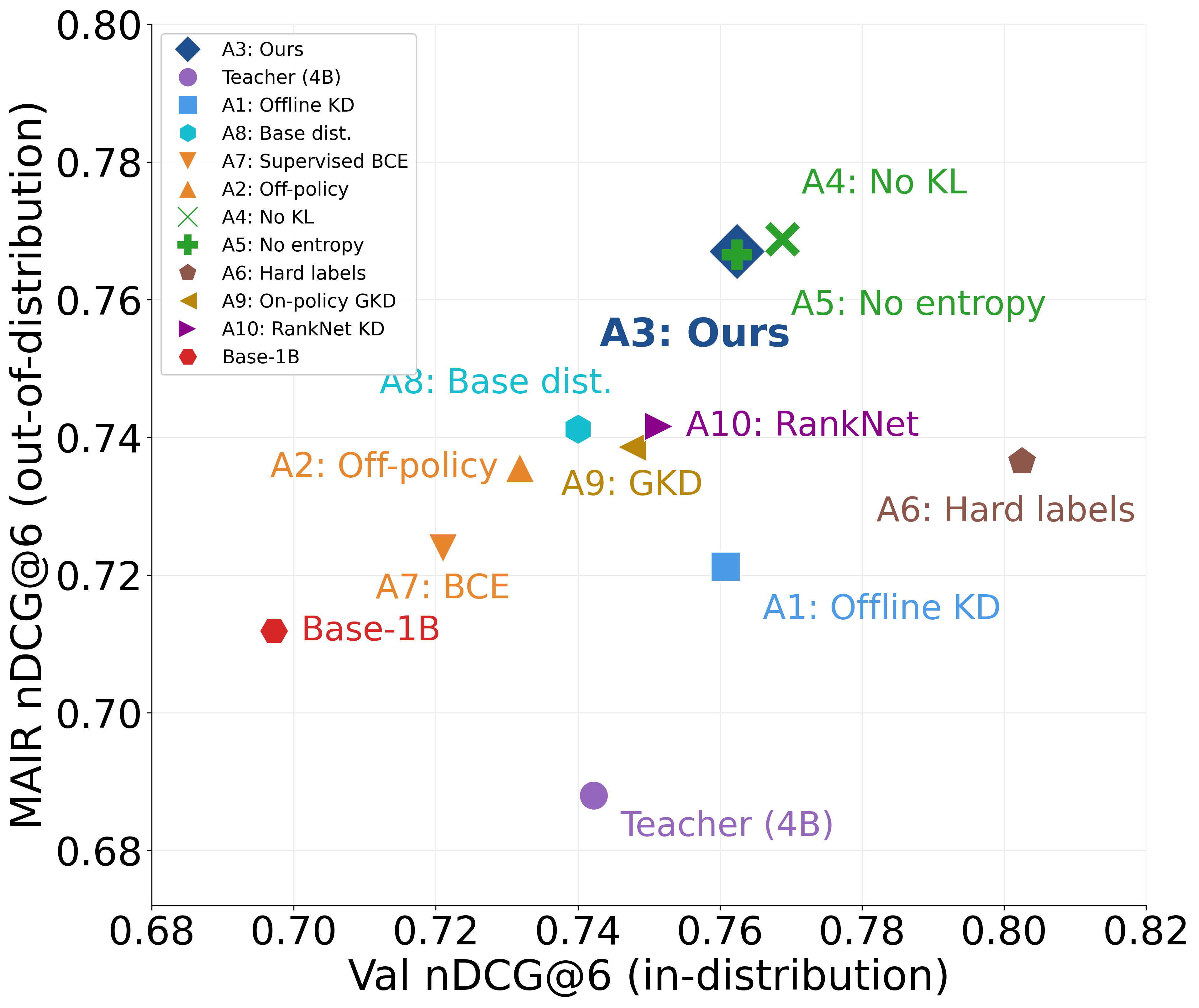}
        \caption{Validation--MAIR-11 tradeoff across core training variants. Top-right is better.}
        \label{fig:val_mair_scatter}
    \end{subfigure}
    \caption{Core ablation evidence for the proposed training pipeline.
    On-policy student-driven distillation provides the strongest overall
    tradeoff between in-distribution fit and OOD generalization, while
    hard-label supervision improves validation performance but harms
    transfer under distribution shift.}
    \label{fig:ablation_overview}
\end{figure*}

Our ablations are designed to answer four primary mechanistic
questions rather than to crown a single best configuration:
(i) does reward-based on-policy distillation outperform alternative
combinations of sampling policy and distillation objective?
(ii) does \emph{where} supervision is evaluated matter when the reward
source is held fixed?
(iii) is Stage~1 teacher strengthening necessary for effective
on-policy distillation?
and (iv) does soft teacher supervision improve OOD robustness relative
to hard labels? We then separately analyze the auxiliary KL and entropy
terms. Figure~\ref{fig:ablation_overview} summarizes the validation
comparison and validation--MAIR-11 tradeoff across the core variants. The analyses below address these questions in turn.

\paragraph{Sampling policy and supervision form.}
A9 and A10 complete a four-way comparison of distillation strategies:
A1 performs offline listwise KD, A10 performs offline pairwise RankNet
KD, A9 performs GKD-style teacher-distribution matching on
student-sampled rankings, and A3 performs reward-based on-policy
distillation over student-sampled ranking permutations. On validation,
A1, A10, A9, and A3 obtain 0.7608, 0.7514, 0.7476, and 0.7624
nDCG@6, respectively. On MAIR-11, they obtain 0.7212, 0.7416,
0.7386, and 0.7670. Thus, neither changing the offline distillation
objective from listwise to pairwise nor performing teacher-distribution
matching on student-sampled rankings reproduces the OOD performance
of reward-based on-policy distillation.

\paragraph{Where supervision is evaluated is the key mechanism.}
Replacing on-policy student sampling with teacher-driven off-policy
sampling (A2) consistently degrades performance, both on the validation
benchmark and under distribution shift. Since A2 and A3 both use
teacher-derived supervision but differ in whose policy generates the
rankings, this gap isolates the role of \emph{student exploration}
rather than reward source alone. These results support our central claim
that effective distillation depends on evaluating supervision on the
student's own sampled ranking space.

\paragraph{Teacher strengthening is a necessary precondition.}
We additionally test whether Stage~1 teacher optimization is necessary
by running on-policy GRPO from the unaligned base ZeRank-2 teacher
(A8). This variant improves over offline KD and the off-policy student,
but remains well below A3 on MAIR-11. On the validation benchmark, A8
remains competitive on several subsets; its main value here is as an
OOD ablation. The result indicates that the gains of the two-stage
pipeline do not come from student-side on-policy learning alone: they
also depend on first improving the teacher with instruction-aware RL
before distillation.

\paragraph{Soft rewards trade in-distribution fit for OOD robustness.}
Replacing soft teacher rewards with hard labels (A6) yields the best
validation performance among the core variants, but substantially
reduces OOD performance on MAIR-11. This contrast exposes a clear
accuracy--generalization tradeoff: hard labels fit the in-distribution
benchmark more aggressively, whereas soft rewards preserve relative
preference information that regularizes the student under distribution
shift. For deployment-oriented instruction-following reranking, this
tradeoff favors soft supervision over maximal validation fit.

\paragraph{Policy-gradient training is the dominant learning signal.}
Removing the KL term (A4) or entropy bonus (A5) leaves performance close
to the full objective, indicating that policy-gradient optimization with
teacher rewards is the primary source of improvement. A5 is numerically very close to A3 on both validation and MAIR-11, while A4 is slightly stronger than A3 both in aggregate and in paired query-level tests. We therefore do not claim that the full objective is uniquely best-performing in every setting. Instead, we view A3 as a robust reference configuration that preserves the complete exploration-aware formulation of the method, while A4 and A5 show that the auxiliary KL and entropy terms are secondary to the main gain from student-driven on-policy distillation with soft teacher rewards.

To complement the aggregate tables, we also compare key variants with
paired Wilcoxon signed-rank tests over query-level scores. The proposed
on-policy student (A3) significantly outperforms offline KD (A1) on
MAIR-11 (two-sided $p=1.95\times10^{-9}$; one-sided A3$>$A1
$p=9.76\times10^{-10}$), strengthening the claim that on-policy
distillation yields substantially more robust ranking behavior under
distribution shift. In contrast, the no-KL variant (A4) achieves
slightly stronger query-level performance than A3 on MAIR-11 (two-sided
$p=0.0269$; one-sided A3$<$A4 $p=0.0134$). However, this aggregate
result does not imply uniform dominance: the subset-level picture
remains mixed, with A3 outperforming A4 on some MAIR-11 subsets and A4
stronger on others. We therefore interpret the core mechanism of the
paper as student-driven on-policy distillation with soft teacher
rewards, while treating the KL term as an optional regularizer rather
than a uniformly beneficial component.

\begin{figure}[t]
\centering
\includegraphics[width=\columnwidth]{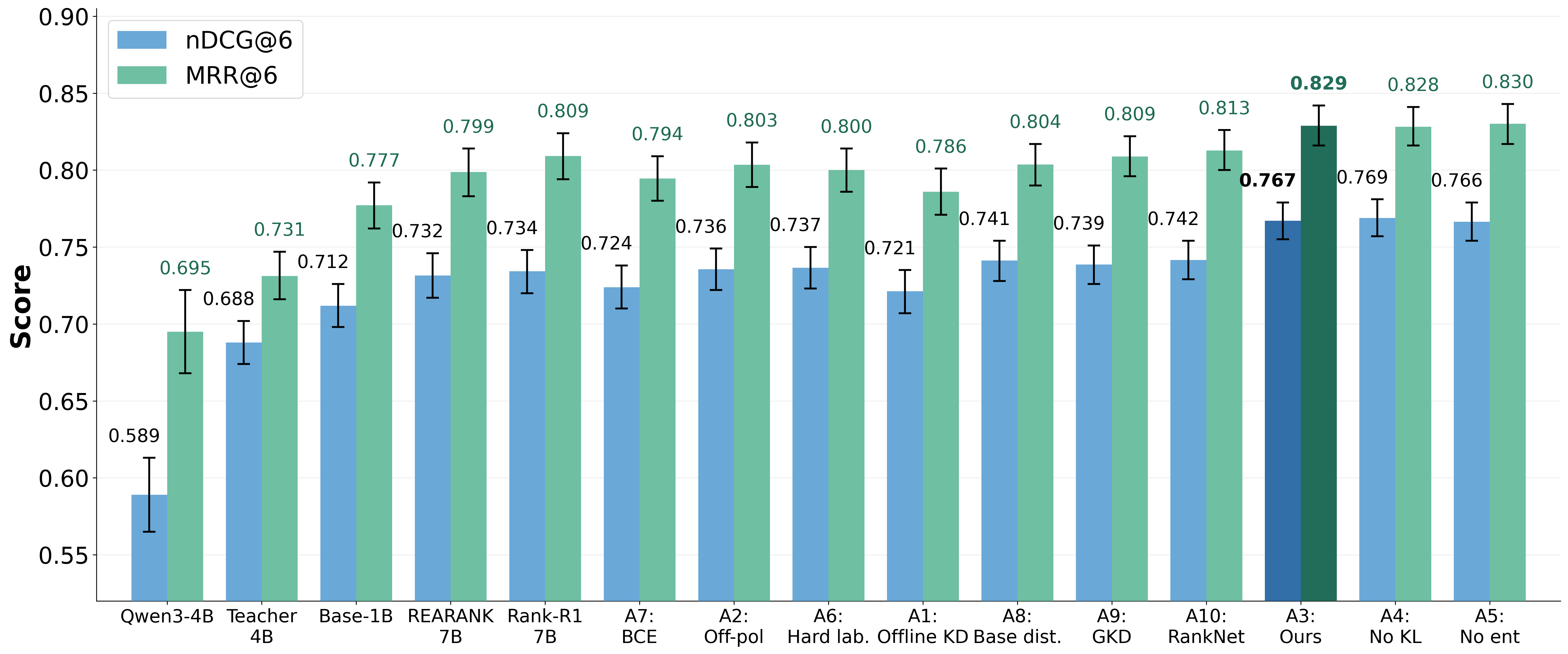}
\caption{MAIR-11 out-of-distribution performance across representative
reference models and core training variants. Error bars indicate
bootstrap 95\% confidence intervals.}
\label{fig:mair_bar}
\end{figure}

To further illustrate out-of-distribution behavior,
Figure~\ref{fig:mair_bar} shows MAIR-11 performance across the core
training variants and representative reference models. The same pattern
observed in Table~\ref{tab:mair_compact} holds visually at both nDCG@6
and MRR@6: on-policy student-driven distillation (A3) substantially
improves over the teacher and the off-policy variant (A2), while
hard-label training (A6) underperforms A3 under distribution shift
despite its stronger in-distribution fit. The figure also makes clear
that the gains are not limited to a single metric, but are consistent
across both ranking quality and first-hit retrieval quality.

\paragraph{The training mechanism transfers across student architectures.}
To test whether the Stage~2 result depends on the
Llama-Nemotron-1B student architecture, we keep the strengthened
Stage~1 ZeRank-2 teacher fixed and repeat Stage~2 training with three
architecturally distinct students: BGE-Reranker-v2-Gemma,
MXBAI-Rerank-Large-v1, and RankZephyr-7B. All three improve after
training. Validation nDCG@6 increases from 0.7015 to 0.7335 for
BGE-Gemma, 0.7016 to 0.7247 for MXBAI, and 0.7132 to 0.7416 for
RankZephyr. On MAIR-11, the corresponding improvements are
0.4866 to 0.8173, 0.6632 to 0.7017, and 0.7024 to 0.7382.
The consistent gains across substantially different student
architectures and reranking interfaces suggest that the Stage~2
training mechanism is not specific to the primary
Llama-Nemotron student. Complete results are reported in
Appendix~\ref{sec:cross_architecture}.

\subsection{Hyperparameter Sensitivity}

We next test the robustness of the default training configuration by
sweeping group size, Plackett--Luce temperature, KL weight, and entropy
coefficient. Across these sweeps, validation performance remains nearly
flat, indicating that the method is not highly sensitive to
hyperparameter tuning.

The no-entropy setting appears both as the Set~A core ablation A5 and as
the Set~B sweep point B4 ($\mathrm{ent}=0.0$); the two are numerically
aligned as expected.

This stability is important for two reasons. First, it shows that the
observed gains are not the result of narrow tuning around a single
configuration. Second, it clarifies the role of the auxiliary terms:
while validation performance changes little across the sweeps, MAIR-11 shows modest sensitivity to the KL coefficient, with stronger KL
regularization improving robustness under distribution shift. This is consistent with the view that KL regularization is not the main driver of learning, but can still improve generalization stability. The full Set~B plot and numeric tables are provided in Appendix~\ref{sec:appendix_setb}.

Overall, the primary benefit comes from \emph{on-policy
policy-gradient training with soft teacher rewards}; KL and entropy
serve as secondary regularizers rather than essential components.

\section{Efficiency Analysis}
\label{sec:efficiency}

A central motivation for compact rerankers is deployment efficiency. Table~\ref{tab:latency} compares single-query inference latency under a fixed hardware setup.

\begin{table}[t]
\centering
\small
\setlength{\tabcolsep}{5pt}
\begin{tabular}{lcccc}
\toprule
\textbf{Model} & \textbf{Params} & \textbf{p50} & \textbf{p95} & \textbf{Mean$\pm$SD} \\
\midrule
Distilled-1B (ours) & 1235.8M & 9.2 & 10.0 & 9.2 $\pm$ 0.3 \\
ZeRank-2 Teacher (4B) & 4022.5M & 24.9 & 41.8 & 27.0 $\pm$ 4.0 \\
BGE Reranker v2 M3 & 567.8M & 7.9 & 16.5 & 9.2 $\pm$ 2.5 \\
Jina Reranker v2 & 278.4M & 5.5 & 18.7 & 7.6 $\pm$ 3.5 \\
Qwen3-Reranker-4B & 4021.8M & 37.9 & 55.0 & 38.9 $\pm$ 4.5 \\
\bottomrule
\end{tabular}
\caption{Single-query latency under a fixed hardware setup.}
\label{tab:latency}
\end{table}

Distilled-1B offers a favorable efficiency--quality tradeoff. It is
substantially faster than 4B rerankers while remaining competitive with
or stronger than all evaluated baselines on ranking quality. Compared to
the teacher, the student reduces mean latency from 27.0ms to 9.2ms
while also improving validation performance.

\section{Discussion and Conclusion}
\label{sec:discussion}

Our results support four conclusions. First, the strongest distillation
behavior arises from combining student-driven sampling with
permutation-level teacher rewards. A9 and A10 show that neither
on-policy teacher-distribution matching nor offline pairwise
distillation reproduces A3's OOD performance, while the A2--A3
comparison isolates the importance of evaluating supervision on the
student's own ranking distribution.

Second, the effect extends beyond the original MAIR-11 evaluation.
Across all 126 MAIR tasks, A3 obtains the highest task-macro point
estimates among the evaluated distillation variants. A3 also exceeds
the evaluated 7B Rank-R1 and REARANK models on the comparable
MAIR-11 setting despite its substantially smaller size.

Third, Stage~1 teacher strengthening and soft reward supervision
remain complementary: distillation from the unaligned teacher (A8)
stays below the full method under distribution shift, while hard-label
training (A6) fits the validation benchmark more aggressively but
generalizes less robustly.

Finally, the Stage~2 mechanism is not tied to the primary
Llama-Nemotron student: the same strengthened teacher and
on-policy procedure improves BGE-Reranker-v2-Gemma,
MXBAI-Rerank-Large-v1, and RankZephyr-7B, three architecturally
distinct students, supporting reward-based on-policy distillation as
a general training strategy across reranker architectures.

\section{Limitations}

Our study focuses on English-language, text-only reranking. Stage~1
relies on LLM-judge reward signals and therefore inherits potential
judge bias and calibration error. Although we validate the judge
against human relevance annotations, this validation contains only
188 examples. Most controlled ablations also use a single random
seed, so small differences among closely clustered objective variants
should be interpreted cautiously; Appendix~\ref{sec:appendix_multiseed}
provides additional multi-seed evidence for A3, A4, and A5.

We evaluate OOD behavior across all 126 MAIR tasks, but the
Stage~1 teacher is not uniformly robust: on MAIR-11 its aggregate
degradation is concentrated in a small number of subsets, particularly
LitSearch and Touche. Cross-architecture experiments show that
Stage~2 transfers to three substantially different student families,
but all experiments use the same strengthened ZeRank-2 teacher.
Repeating the computationally expensive Stage~1 optimization across
multiple teacher families would require additional multi-GPU training
runs and was outside our available compute budget. Our experiments
therefore establish student-side architectural generality, while
teacher-side architectural and capability generality remain open.

\section{Ethical Considerations}

This work studies instruction-following reranking for text retrieval and
does not introduce new user-facing generation capabilities. The main
ethical risks arise from misranking rather than content generation:
instruction-following rerankers may amplify bias present in training
data, judge models, or candidate corpora, and may under-rank relevant
documents for particular domains, perspectives, or user groups. Because
our Stage~1 teacher is trained with judge-derived reward signals, the
pipeline can also inherit calibration errors or systematic preferences
from the judge model. We partially mitigate this risk by validating the
judge against both other strong LLM evaluators and human relevance
annotations, but these checks do not eliminate all bias.

Our experiments are limited to English text-only reranking and to the
datasets and MAIR benchmark evaluated in the paper, so the findings
should not be interpreted as guaranteeing fair or robust behavior
across all domains or deployment settings. In enterprise use, instruction-following rerankers should be monitored with domain-appropriate evaluation, auditing, and human oversight, especially in high-stakes settings such as legal, medical, financial, or employment-related retrieval.

\section*{Acknowledgments}
This work was supported by SAP, and all training and evaluation were
conducted on SAP-provided NVIDIA H200 GPU infrastructure. We thank our colleagues on the SAP Business AI team for helpful discussions and feedback as well as the anonymous reviewers and meta-reviewer for suggestions that substantially improved the paper. 

Generative AI assistants were used for language editing, manuscript consistency checks and limited coding/debugging assistance. All scientific decisions, experiments, analyses, citations, interpretations and final manuscript content were reviewed and verified by the authors.


\bibliography{custom}

@article{nogueira2019passage,
  title   = {Passage Re-ranking with {BERT}},
  author  = {Nogueira, Rodrigo and Cho, Kyunghyun},
  journal = {arXiv preprint arXiv:1901.04085},
  year    = {2019},
  url     = {https://arxiv.org/abs/1901.04085}
}

@article{khattab2020colbert,
  title   = {{ColBERT}: Efficient and Effective Passage Search via Contextualized Late Interaction over {BERT}},
  author  = {Khattab, Omar and Zaharia, Matei},
  journal = {arXiv preprint arXiv:2004.12832},
  year    = {2020},
  url     = {https://arxiv.org/abs/2004.12832}
}

@inproceedings{lin2021pretrained,
  title     = {Pretrained Transformers for Text Ranking: {BERT} and Beyond},
  author    = {Yates, Andrew and Nogueira, Rodrigo and Lin, Jimmy},
  booktitle = {Proceedings of the 2021 Conference of the North American Chapter of the Association for Computational Linguistics: Human Language Technologies: Tutorials},
  pages     = {1--4},
  year      = {2021},
  address   = {Online},
  publisher = {Association for Computational Linguistics},
  doi       = {10.18653/v1/2021.naacl-tutorials.1},
  url       = {https://aclanthology.org/2021.naacl-tutorials.1/}
}

@inproceedings{asai2023task,
  title     = {Task-aware Retrieval with Instructions},
  author    = {Asai, Akari and Schick, Timo and Lewis, Patrick and Chen, Xilun and Izacard, Gautier and Riedel, Sebastian and Hajishirzi, Hannaneh and Yih, Wen-tau},
  booktitle = {Findings of the Association for Computational Linguistics: ACL 2023},
  pages     = {3650--3675},
  year      = {2023},
  address   = {Toronto, Canada},
  publisher = {Association for Computational Linguistics},
  doi       = {10.18653/v1/2023.findings-acl.225},
  url       = {https://aclanthology.org/2023.findings-acl.225/}
}

@inproceedings{weller2024followir,
  title     = {{FollowIR}: Evaluating and Teaching Information Retrieval Models to Follow Instructions},
  author    = {Weller, Orion and Chang, Benjamin and MacAvaney, Sean and Lo, Kyle and Cohan, Arman and Van Durme, Benjamin and Lawrie, Dawn and Soldaini, Luca},
  booktitle = {Proceedings of the 2025 Conference of the Nations of the Americas Chapter of the Association for Computational Linguistics: Human Language Technologies (Volume 1: Long Papers)},
  pages     = {11926--11942},
  year      = {2025},
  address   = {Albuquerque, New Mexico},
  publisher = {Association for Computational Linguistics},
  doi       = {10.18653/v1/2025.naacl-long.597},
  url       = {https://aclanthology.org/2025.naacl-long.597/}
}

@article{oh2024instructir,
  title   = {{INSTRUCTIR}: A Benchmark for Instruction Following of Information Retrieval Models},
  author  = {Oh, Hanseok and Lee, Hyunji and Ye, Seonghyeon and Shin, Haebin and Jang, Hansol and Jun, Changwook and Seo, Minjoon},
  journal = {arXiv preprint arXiv:2402.14334},
  year    = {2024},
  url     = {https://arxiv.org/abs/2402.14334}
}

@inproceedings{ouyang2022training,
  title     = {Training Language Models to Follow Instructions with Human Feedback},
  author    = {Ouyang, Long and Wu, Jeffrey and Jiang, Xu and Almeida, Diogo and Wainwright, Carroll and Mishkin, Pamela and Zhang, Chong and Agarwal, Sandhini and Slama, Katarina and Ray, Alex and Schulman, John and Hilton, Jacob and Kelton, Fraser and Miller, Luke and Simens, Maddie and Askell, Amanda and Welinder, Peter and Christiano, Paul F. and Leike, Jan and Lowe, Ryan},
  booktitle = {Advances in Neural Information Processing Systems},
  volume    = {35},
  pages     = {27730--27744},
  year      = {2022},
  url       = {https://proceedings.neurips.cc/paper_files/paper/2022/hash/b1efde53be364a73914f58805a001731-Abstract-Conference.html}
}

@inproceedings{rafailov2023direct,
  title     = {Direct Preference Optimization: Your Language Model is Secretly a Reward Model},
  author    = {Rafailov, Rafael and Sharma, Archit and Mitchell, Eric and Manning, Christopher D. and Ermon, Stefano and Finn, Chelsea},
  booktitle = {Advances in Neural Information Processing Systems},
  volume    = {36},
  pages     = {53728--53741},
  year      = {2023},
  url       = {https://proceedings.neurips.cc/paper_files/paper/2023/hash/a85b405ed65c6477a4fe8302b5e06ce7-Abstract-Conference.html}
}

@article{shao2024deepseekmath,
  title   = {{DeepSeekMath}: Pushing the Limits of Mathematical Reasoning in Open Language Models},
  author  = {Shao, Zhihong and Wang, Peiyi and Zhu, Qihao and Xu, Runxin and Song, Junxiao and Bi, Xiao and Zhang, Haowei and Zhang, Mingchuan and Li, Y. K. and Wu, Y. and Guo, Daya},
  journal = {arXiv preprint arXiv:2402.03300},
  year    = {2024},
  url     = {https://arxiv.org/abs/2402.03300}
}

@article{hinton2015distilling,
  title   = {Distilling the Knowledge in a Neural Network},
  author  = {Hinton, Geoffrey and Vinyals, Oriol and Dean, Jeff},
  journal = {arXiv preprint arXiv:1503.02531},
  year    = {2015},
  url     = {https://arxiv.org/abs/1503.02531}
}

@inproceedings{hofstatter2021efficiently,
  title     = {Efficiently Teaching an Effective Dense Retriever with Balanced Topic Aware Sampling},
  author    = {Hofst{\"a}tter, Sebastian and Lin, Sheng-Chieh and Yang, Jheng-Hong and Lin, Jimmy and Hanbury, Allan},
  booktitle = {Proceedings of the 44th International ACM SIGIR Conference on Research and Development in Information Retrieval},
  pages     = {113--122},
  year      = {2021},
  publisher = {Association for Computing Machinery},
  doi       = {10.1145/3404835.3462891},
  url       = {https://doi.org/10.1145/3404835.3462891}
}

@inproceedings{furlanello2018born,
  title     = {Born Again Neural Networks},
  author    = {Furlanello, Tommaso and Lipton, Zachary C. and Tschannen, Michael and Itti, Laurent and Anandkumar, Anima},
  booktitle = {Proceedings of the 35th International Conference on Machine Learning},
  series    = {Proceedings of Machine Learning Research},
  volume    = {80},
  pages     = {1607--1616},
  year      = {2018},
  publisher = {PMLR},
  url       = {https://proceedings.mlr.press/v80/furlanello18a.html}
}

@inproceedings{zheng2023judging,
  title     = {Judging {LLM}-as-a-Judge with {MT-Bench} and Chatbot Arena},
  author    = {Zheng, Lianmin and Chiang, Wei-Lin and Sheng, Ying and Zhuang, Siyuan and Wu, Zhanghao and Zhuang, Yonghao and Lin, Zi and Li, Zhuohan and Li, Dacheng and Xing, Eric P. and Zhang, Hao and Gonzalez, Joseph E. and Stoica, Ion},
  booktitle = {Advances in Neural Information Processing Systems},
  volume    = {36},
  pages     = {46595--46623},
  year      = {2023},
  publisher = {Curran Associates, Inc.},
  url       = {https://proceedings.neurips.cc/paper_files/paper/2023/hash/91f18a1287b398d378ef22505bf41832-Abstract-Datasets_and_Benchmarks.html}
}

@article{dubois2024alpacaeval,
  title   = {Length-Controlled AlpacaEval: A Simple Way to Debias Automatic Evaluators},
  author  = {Dubois, Yann and Galambosi, Bal{\'a}zs and Liang, Percy and Hashimoto, Tatsunori B.},
  journal = {arXiv preprint arXiv:2404.04475},
  year    = {2024},
  url     = {https://arxiv.org/abs/2404.04475}
}

@misc{zerank2model,
  title        = {zeroentropy/zerank-2-reranker},
  author       = {{ZeroEntropy}},
  year         = {2025},
  howpublished = {\url{https://huggingface.co/zeroentropy/zerank-2-reranker}},
  note         = {Hugging Face model card, 4B instruction-following reranker},
  url          = {https://huggingface.co/zeroentropy/zerank-2-reranker}
}

@misc{nemotronrerank1bv2,
  title        = {nvidia/llama-nemotron-rerank-1b-v2},
  author       = {{NVIDIA}},
  howpublished = {Hugging Face model card},
  year         = {2025},
  note         = {Accessed 2026-04-10},
  url          = {https://huggingface.co/nvidia/llama-nemotron-rerank-1b-v2}
}

@article{jarvelin2002cumulated,
  title   = {Cumulated Gain-based Evaluation of IR Techniques},
  author  = {J{\"a}rvelin, Kalervo and Kek{\"a}l{\"a}inen, Jaana},
  journal = {ACM Transactions on Information Systems},
  volume  = {20},
  number  = {4},
  pages   = {422--446},
  year    = {2002},
  doi     = {10.1145/582415.582418},
  url     = {https://doi.org/10.1145/582415.582418}
}

@inproceedings{chapelle2009expected,
  title     = {Expected Reciprocal Rank for Graded Relevance},
  author    = {Chapelle, Olivier and Metzler, Donald and Zhang, Ya and Grinspan, Pierre},
  booktitle = {Proceedings of the 18th ACM Conference on Information and Knowledge Management},
  pages     = {621--630},
  year      = {2009},
  publisher = {Association for Computing Machinery},
  doi       = {10.1145/1645953.1646033},
  url       = {https://doi.org/10.1145/1645953.1646033}
}

@article{ma2023zeroshotlistwise,
  title   = {Zero-Shot Listwise Document Reranking with a Large Language Model},
  author  = {Ma, Xueguang and Zhang, Xinyu and Pradeep, Ronak and Lin, Jimmy},
  journal = {arXiv preprint arXiv:2305.02156},
  year    = {2023},
  url     = {https://arxiv.org/abs/2305.02156}
}

@article{pradeep2023rankzephyr,
  title   = {RankZephyr: Effective and Robust Zero-Shot Listwise Reranking is a Breeze!},
  author  = {Pradeep, Ronak and Sharifymoghaddam, Sahel and Lin, Jimmy},
  journal = {arXiv preprint arXiv:2312.02724},
  year    = {2023},
  url     = {https://arxiv.org/abs/2312.02724}
}

@inproceedings{weller2024promptriever,
  title     = {Promptriever: Instruction-Trained Retrievers Can Be Prompted Like Language Models},
  author    = {Weller, Orion and Van Durme, Benjamin and Lawrie, Dawn and Paranjape, Ashwin and Zhang, Yuhao and Hessel, Jack},
  booktitle = {International Conference on Learning Representations},
  year      = {2025},
  url       = {https://proceedings.iclr.cc/paper_files/paper/2025/hash/2cefdb2c4c3274b78cd450bac35228df-Abstract-Conference.html}
}

@inproceedings{qin2024rankgpt,
  title     = {Large Language Models are Effective Text Rankers with Pairwise Ranking Prompting},
  author    = {Qin, Zhen and Jagerman, Rolf and Hui, Kai and Zhuang, Honglei and Wu, Junru and Yan, Le and Shen, Jiaming and Liu, Tianqi and Liu, Jialu and Metzler, Donald and Wang, Xuanhui and Bendersky, Michael},
  booktitle = {Findings of the Association for Computational Linguistics: NAACL 2024},
  pages     = {1504--1518},
  year      = {2024},
  address   = {Mexico City, Mexico},
  publisher = {Association for Computational Linguistics},
  doi       = {10.18653/v1/2024.findings-naacl.97},
  url       = {https://aclanthology.org/2024.findings-naacl.97/}
}

@article{pradeep2023rankvicuna,
  title   = {RankVicuna: Zero-Shot Listwise Document Reranking with Open-Source Large Language Models},
  author  = {Pradeep, Ronak and Sharifymoghaddam, Sahel and Lin, Jimmy},
  journal = {arXiv preprint arXiv:2309.15088},
  year    = {2023},
  url     = {https://arxiv.org/abs/2309.15088}
}

@article{levine2020offline,
  title   = {Offline Reinforcement Learning: Tutorial, Review, and Perspectives on Open Problems},
  author  = {Levine, Sergey and Kumar, Aviral and Tucker, George and Fu, Justin},
  journal = {arXiv preprint arXiv:2005.01643},
  year    = {2020},
  url     = {https://arxiv.org/abs/2005.01643}
}

@inproceedings{agarwal2024onpolicy,
  title     = {On-Policy Distillation of Language Models: Learning from Self-Generated Mistakes},
  author    = {Agarwal, Rishabh and Vieillard, Nino and Zhou, Yongchao and Stanczyk, Piotr and Ramos Garea, Sabela and Geist, Matthieu and Bachem, Olivier},
  booktitle = {International Conference on Learning Representations},
  year      = {2024},
  url       = {https://proceedings.iclr.cc/paper_files/paper/2024/hash/5be69a584901a26c521c2b51e40a4c20-Abstract-Conference.html}
}

@article{lu2025onpolicydistillation,
  author  = {Lu, Kevin and {Thinking Machines Lab}},
  title   = {On-Policy Distillation},
  journal = {Thinking Machines Lab: Connectionism},
  year    = {2025},
  doi     = {10.64434/tml.20251026},
  url     = {https://thinkingmachines.ai/blog/on-policy-distillation/}
}

@article{qwen3,
  title   = {Qwen3 Technical Report},
  author  = {{Qwen Team}},
  journal = {arXiv preprint arXiv:2505.09388},
  year    = {2025},
  url     = {https://arxiv.org/abs/2505.09388}
}

@inproceedings{zhuang2026rankr1,
  title     = {Rank-R1: Enhancing Reasoning in LLM-based Document Rerankers via Reinforcement Learning},
  author    = {Zhuang, Shengyao and Ma, Xueguang and Yao, Zheng and Wang, Shuai and Koopman, Bevan and Lin, Jimmy and Zuccon, Guido},
  booktitle = {Proceedings of the 49th International ACM SIGIR Conference on Research and Development in Information Retrieval},
  year      = {2026},
  pages     = {4419--4425},
  publisher = {Association for Computing Machinery},
  doi       = {10.1145/3805712.3809961},
  url       = {https://doi.org/10.1145/3805712.3809961}
}

@inproceedings{zhang2025rearank,
  title     = {{REARANK}: Reasoning Re-ranking Agent via Reinforcement Learning},
  author    = {Zhang, Le and Wang, Bo and Qiu, Xipeng and Reddy, Siva and Agrawal, Aishwarya},
  booktitle = {Proceedings of the 2025 Conference on Empirical Methods in Natural Language Processing},
  year      = {2025},
  pages     = {2458--2471},
  address   = {Suzhou, China},
  publisher = {Association for Computational Linguistics},
  doi       = {10.18653/v1/2025.emnlp-main.125},
  url       = {https://aclanthology.org/2025.emnlp-main.125/}
}

@inproceedings{sun2024mair,
  title     = {{MAIR}: A Massive Benchmark for Evaluating Instructed Retrieval},
  author    = {Sun, Weiwei and Shi, Zhengliang and Long, Wu Jiu and Yan, Lingyong and Ma, Xinyu and Liu, Yiding and Cao, Min and Yin, Dawei and Ren, Zhaochun},
  booktitle = {Proceedings of the 2024 Conference on Empirical Methods in Natural Language Processing},
  year      = {2024},
  pages     = {14044--14067},
  address   = {Miami, Florida, USA},
  publisher = {Association for Computational Linguistics},
  doi       = {10.18653/v1/2024.emnlp-main.778},
  url       = {https://aclanthology.org/2024.emnlp-main.778/}
}

@inproceedings{choi2024rradistill,
  title     = {{RRADistill}: Distilling LLMs' Passage Ranking Ability for Long-Tail Queries Document Re-Ranking on a Search Engine},
  author    = {Choi, Nayoung and Lee, Youngjune and Cho, Gyu-Hwung and Jeong, Haeyu and Kong, Jungmin and Kim, Saehun and Park, Keunchan and Cho, Sarah and Jeong, Inchang and Nam, Gyohee and Han, Sunghoon and Yang, Wonil and Choi, Jaeho},
  booktitle = {Proceedings of the 2024 Conference on Empirical Methods in Natural Language Processing: Industry Track},
  year      = {2024},
  pages     = {627--641},
  address   = {Miami, Florida, US},
  publisher = {Association for Computational Linguistics},
  doi       = {10.18653/v1/2024.emnlp-industry.46},
  url       = {https://aclanthology.org/2024.emnlp-industry.46/}
}

@inproceedings{huang2025gumbel,
  title     = {Gumbel Reranking: Differentiable End-to-End Reranker Optimization},
  author    = {Huang, Siyuan and Ma, Zhiyuan and Du, Jintao and Meng, Changhua and Wang, Weiqiang and Leng, Jingwen and Guo, Minyi and Lin, Zhouhan},
  booktitle = {Proceedings of the 63rd Annual Meeting of the Association for Computational Linguistics (Volume 1: Long Papers)},
  year      = {2025},
  pages     = {7142--7161},
  address   = {Vienna, Austria},
  publisher = {Association for Computational Linguistics},
  doi       = {10.18653/v1/2025.acl-long.354},
  url       = {https://aclanthology.org/2025.acl-long.354/}
}

@inproceedings{xu2025distillation,
  title     = {Distillation versus Contrastive Learning: How to Train Your Rerankers},
  author    = {Xu, Zhichao and Huang, Zhiqi and Zhuang, Shengyao and Srikumar, Vivek},
  booktitle = {Proceedings of the 14th International Joint Conference on Natural Language Processing and the 4th Conference of the Asia-Pacific Chapter of the Association for Computational Linguistics},
  year      = {2025},
  pages     = {564--578},
  address   = {Mumbai, India},
  publisher = {The Asian Federation of Natural Language Processing and The Association for Computational Linguistics},
  doi       = {10.18653/v1/2025.findings-ijcnlp.33},
  url       = {https://aclanthology.org/2025.findings-ijcnlp.33/}
}

@inproceedings{cai2026erank,
  title     = {{ERank}: Fusing Supervised Fine-Tuning and Reinforcement Learning for Effective and Efficient Text Reranking},
  author    = {Cai, Yuzheng and Zhang, Yanzhao and Long, Dingkun and Li, Mingxin and Xie, Pengjun and Zheng, Weiguo},
  booktitle = {Proceedings of the AAAI Conference on Artificial Intelligence},
  volume    = {40},
  number    = {36},
  year      = {2026},
  pages     = {30121--30129},
  doi       = {10.1609/aaai.v40i36.40261},
  url       = {https://ojs.aaai.org/index.php/AAAI/article/view/40261}
}

\clearpage
\appendix

\section{Reproducibility and Compute}
\label{sec:appendix_repro}
To support reproducibility, we release the final Distilled-1B
checkpoint, all datasets used in the paper, and the training and
evaluation code for the main model. All repositories listed below are publicly accessible and require no authentication to download.

\begin{itemize}[leftmargin=*, nosep]
  \item \textbf{Model checkpoint:}
    \url{https://huggingface.co/anonymousauthor01/instruction_following_reranker}
  \item \textbf{Train-validation sets:}
    \url{https://huggingface.co/datasets/anonymousauthor01/emnlp-2026-ifr-train-val-set}
  \item \textbf{MAIR-11 subsets:}
    \url{https://huggingface.co/datasets/anonymousauthor01/emnlp-2026-ifr-mair-ood}
  \item \textbf{MAIR-Full dataset:}
    \url{https://huggingface.co/datasets/anonymousauthor01/emnlp-2026-ifr-mair-full}
  \item \textbf{Code repository with instructions:}
    \url{https://github.com/vigneshprabhakar1998/emnlp-2026-artifact-release}
\end{itemize}

\noindent The code repository contains the
training scripts, the evaluation scripts, their corresponding
launchers, and the exact training and evaluation configuration files
used for the reported A3 runs, including the metric and
bootstrap confidence-interval implementations. The HuggingFace
repositories provide the final A3 checkpoint together with the three
preprocessed datasets used in the paper (the instruction-following train/validation set, MAIR-11, and MAIR-Full). Together these reproduce the reported validation, MAIR-11, and MAIR-Full results for the final model using the same candidate sets, relevance labels, metrics, and bootstrap confidence-interval procedure used in the paper.

All reported training and evaluation runs were conducted on NVIDIA H200
GPUs. The Stage~1 teacher run required approximately 3 days and 23 hours
on 2 H200 GPUs, the default Stage~2 on-policy distillation run required
approximately 4 hours and 50 minutes on a single H200, and the full
validation+MAIR evaluation suite required approximately 20 minutes on a
single GPU. The core reported pipeline therefore required approximately
195 H200 GPU-hours in total, excluding non-reported debugging and
exploratory runs.

\section{Additional Experimental Details}
\label{sec:appendix_overview}

This appendix provides supplementary evidence for the main empirical
claims of the paper. In addition to the compact results presented in the
main text, we report: (i) the complete MAIR-Full evaluation across all
126 tasks and 9,356 queries, together with results on the 115 tasks not
included in MAIR-11; (ii) detailed MAIR-11 and validation ablation
tables, including the A9 on-policy GKD and A10 RankNet pairwise KD
controls; (iii) cross-architecture transfer experiments using
BGE-Reranker-v2-Gemma, MXBAI-Rerank-Large-v1, and RankZephyr-7B;
(iv) the full Set~B hyperparameter sweeps; and (v) subset-level,
statistical, multi-seed, and qualitative analyses that further
characterize the behavior of the proposed method.

Across these analyses, the central pattern is consistent: the strongest
OOD behavior comes from combining student-driven sampling with
permutation-level teacher rewards. Neither offline pairwise
distillation nor on-policy teacher-distribution matching reproduces the
MAIR-11 performance of A3, and A3 remains the strongest evaluated
distillation variant when evaluation is expanded to MAIR-Full. At the
same time, the appendix qualifies the aggregate story. Closely related
objective variants such as A4 and A5 remain tightly clustered with A3
and can outperform it on individual subsets or metrics, while the
Stage~1 teacher itself exhibits concentrated rather than uniform OOD
failures. The cross-architecture results further show that the Stage~2
training mechanism transfers across substantially different student
backbones, while teacher-side architectural generality remains outside
the scope of the present experiments.

\section{MAIR-Full Evaluation}
\label{sec:appendix_mair_full126}

Table~\ref{tab:mair_full126} reports task-macro performance across
the complete MAIR benchmark of 126 tasks and 9,356 queries. A3
obtains the highest point estimate for both nDCG@6 and MRR@6 among
the evaluated distillation variants, followed by A9. This extends the
main MAIR-11 observation to a substantially broader collection of
retrieval tasks.

\begin{table*}[t]
\centering
\small
\setlength{\tabcolsep}{11pt}
\begin{tabular}{lllcc}
\toprule
\textbf{Rank} & \textbf{Variant} & \textbf{Method} &
\textbf{nDCG@6 [95\% CI]} &
\textbf{MRR@6 [95\% CI]} \\
\midrule
1 & \textbf{A3} & \textbf{On-policy GRPO distillation (ours)}
& \textbf{0.6808 [0.6741, 0.7099]}
& \textbf{0.7865 [0.7669, 0.8010]} \\

2 & A9 & On-policy GKD
& 0.6657 [0.6475, 0.6725]
& 0.7565 [0.7323, 0.7650] \\

3 & A1 & Offline listwise KD
& 0.6564 [0.6136, 0.6691]
& 0.7512 [0.7356, 0.7597] \\

4 & A10 & RankNet pairwise KD
& 0.6521 [0.6142, 0.6662]
& 0.7502 [0.7338, 0.7590] \\

5 & A6 & Hard-label distillation
& 0.6475 [0.6075, 0.6612]
& 0.7286 [0.6928, 0.7533] \\

6 & A2 & Off-policy GRPO distillation
& 0.6438 [0.6058, 0.6567]
& 0.7438 [0.7009, 0.7506] \\
\bottomrule
\end{tabular}
\caption{MAIR-Full evaluation across 126 tasks and 9,356 queries.
Values are task-macro means with 95\% bootstrap confidence intervals
from 10,000 resamples.}
\label{tab:mair_full126}
\end{table*}

\subsection{Performance on the Additional 115 MAIR Tasks}

To isolate the newly added evaluation coverage, Table~\ref{tab:mair_115}
reports task-macro performance on the 115 MAIR tasks not included in
MAIR-11, comprising 8,487 queries. A3 again obtains the highest
point estimate for both ranking metrics.

\begin{table*}[t]
\centering
\small
\setlength{\tabcolsep}{10pt}
\begin{tabular}{llcc}
\toprule
\textbf{Variant} & \textbf{Method} &
\textbf{nDCG@6 [95\% CI]} &
\textbf{MRR@6 [95\% CI]} \\
\midrule
A1 & Offline listwise KD
& 0.6498 [0.6031, 0.6624]
& 0.7476 [0.7312, 0.7562] \\

A2 & Off-policy GRPO distillation
& 0.6344 [0.6051, 0.6462]
& 0.7377 [0.6929, 0.7428] \\

\textbf{A3} & \textbf{On-policy GRPO distillation}
& \textbf{0.6720 [0.6668, 0.7018]}
& \textbf{0.7821 [0.7628, 0.7959]} \\

A6 & Hard-label distillation
& 0.6384 [0.5967, 0.6512]
& 0.7213 [0.6842, 0.7461] \\

A9 & On-policy GKD
& 0.6582 [0.6393, 0.6652]
& 0.7512 [0.7260, 0.7602] \\

A10 & RankNet pairwise KD
& 0.6429 [0.6019, 0.6582]
& 0.7438 [0.7262, 0.7528] \\
\bottomrule
\end{tabular}
\caption{Task-macro results on the 115 MAIR tasks not included in
MAIR-11, comprising 8,487 queries. Values are means with 95\%
bootstrap confidence intervals.}
\label{tab:mair_115}
\end{table*}

\section{Detailed MAIR-11 Ablation Results}
\label{sec:appendix_mair}

Table~\ref{tab:appendix_mair_full} reports the complete controlled
ablation results on MAIR-11. This view complements the broader
126-task evaluation by retaining the original 869-query setting used
for direct comparison among all core Set~A variants.

\begin{table*}[!t]
\centering
\small
\setlength{\tabcolsep}{7pt}
\begin{tabular}{lcccc}
\toprule
\textbf{Model} &
\textbf{nDCG@6 [95\% CI]} & \textbf{SD} &
\textbf{MRR@6 [95\% CI]} & \textbf{SD} \\
\midrule
Base-1B (no training)
& 0.7119 [0.698, 0.726] & 0.2150
& 0.7771 [0.762, 0.792] & 0.2250 \\
ZeRank-2 base (4B)
& 0.6938 [0.679, 0.708] & 0.2140
& 0.7266 [0.711, 0.742] & 0.2320 \\
Teacher GRPO (4B)
& 0.6880 [0.674, 0.702] & 0.2150
& 0.7312 [0.716, 0.747] & 0.2350 \\
A7: Supervised BCE
& 0.7239 [0.710, 0.738] & 0.2070
& 0.7945 [0.780, 0.809] & 0.2180 \\
A1: Offline KD
& 0.7212 [0.707, 0.735] & 0.2080
& 0.7860 [0.771, 0.801] & 0.2200 \\
A2: Off-policy GRPO
& 0.7356 [0.722, 0.749] & 0.2050
& 0.8035 [0.789, 0.818] & 0.2150 \\
A8: On-policy GRPO from base ZeRank-2
& 0.7412 [0.728, 0.754] & 0.1980
& 0.8036 [0.790, 0.817] & 0.2050 \\
\textbf{A3: On-policy GRPO from Teacher (ours)}
& 0.7670 [0.755, 0.779] & 0.1850
& 0.8289 [0.816, 0.842] & 0.1900 \\
A4: No KL term
& 0.7688 [0.757, 0.781] & 0.1820
& 0.8282 [0.816, 0.841] & 0.1880 \\
A5: No entropy
& 0.7665 [0.754, 0.779] & 0.1860
& 0.8301 [0.817, 0.843] & 0.1900 \\
A6: Hard labels
& 0.7365 [0.723, 0.750] & 0.2080
& 0.8000 [0.786, 0.814] & 0.2150 \\
A9: On-policy GKD
& 0.7386 [0.726, 0.751] & 0.1920
& 0.8088 [0.796, 0.822] & 0.1950 \\
A10: RankNet pairwise KD
& 0.7416 [0.729, 0.754] & 0.1900
& 0.8128 [0.800, 0.826] & 0.1920 \\
\bottomrule
\end{tabular}
\caption{Detailed MAIR-11 ablation results on 869 queries.
Confidence intervals are bootstrap 95\% CIs with 10,000 resamples.}
\label{tab:appendix_mair_full}
\end{table*}

\subsection{Comparable Distillation Ablations on Validation}

Table~\ref{tab:appendix_distillation_val} compares the five
distillation strategies for which directly comparable validation
statistics are available. A1 and A3 remain close in-distribution,
whereas the separation among the methods becomes substantially larger
on MAIR-11.

\begin{table*}[t]
\centering
\small
\setlength{\tabcolsep}{8pt}
\begin{tabular}{llcccc}
\toprule
\textbf{Variant} & \textbf{Method} &
\textbf{nDCG@6 [95\% CI]} & \textbf{SD} &
\textbf{MRR@6 [95\% CI]} & \textbf{SD} \\
\midrule
A1 & Offline listwise KD
& 0.7608 [0.7535, 0.7681] & 0.3730
& 0.7460 [0.7383, 0.7537] & 0.3880 \\
A2 & Off-policy GRPO
& 0.7318 [0.7245, 0.7391] & 0.3700
& 0.7100 [0.7023, 0.7177] & 0.3910 \\
\textbf{A3} & \textbf{On-policy GRPO (ours)}
& \textbf{0.7624 [0.7551, 0.7697]} & 0.3730
& \textbf{0.7475 [0.7398, 0.7552]} & 0.3885 \\
A9 & On-policy GKD
& 0.7476 [0.7402, 0.7550] & 0.3733
& 0.7340 [0.7263, 0.7417] & 0.3878 \\
A10 & RankNet pairwise KD
& 0.7514 [0.7441, 0.7587] & 0.3730
& 0.7330 [0.7253, 0.7407] & 0.3879 \\
\bottomrule
\end{tabular}
\caption{Comparable validation results for the primary distillation
strategies on 9,861 queries.}
\label{tab:appendix_distillation_val}
\end{table*}

\subsection{Why does Teacher GRPO underperform Base-1B on MAIR-11?}
\label{sec:appendix_teacher_mechanism}

\begin{figure*}[t]
\centering
\includegraphics[width=0.9\textwidth]{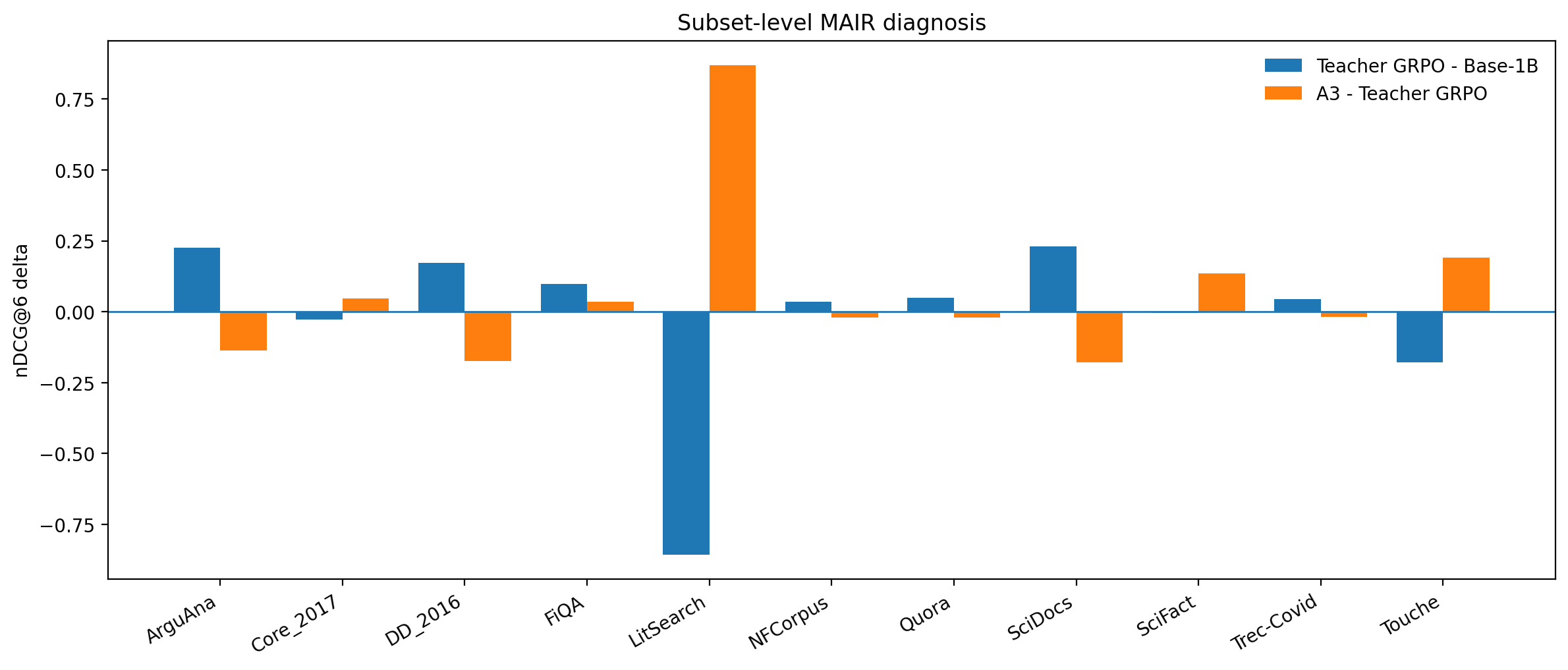}
\caption{Subset-level MAIR-11 diagnosis using nDCG@6 deltas. The left bar
for each subset shows the change from Base-1B to Teacher GRPO; the right
bar shows the change from Teacher GRPO to A3. Teacher GRPO improves over
Base-1B on most subsets, but suffers a few concentrated OOD failures,
especially on LitSearch and Touche. A3's aggregate gain comes mainly
from repairing those largest teacher failures rather than uniformly
dominating the teacher on every subset.}
\label{fig:appendix_teacher_base_a3_mair}
\end{figure*}

The aggregate MAIR-11 results might suggest that the Stage~1 teacher
(\textbf{Teacher GRPO}) simply transfers worse than the untrained
\textbf{Base-1B} model under distribution shift. A subset-level analysis
(Table~\ref{tab:appendix_teacher_base_a3_mair};
Figure~\ref{fig:appendix_teacher_base_a3_mair}) shows a more specific
pattern. Teacher GRPO does \emph{not} degrade
uniformly across MAIR-11: relative to Base-1B, it is stronger on 7 of 11
subsets, including ArguAna, DD\_2016, FiQA, SciDocs, NFCorpus, Quora,
and Trec-Covid. However, this improvement is offset by a small number of
large OOD failures, most notably on LitSearch and Touche, which dominate
the aggregate average.

This pattern suggests that Stage~1 off-policy RL does not merely make the
teacher weaker OOD; rather, it produces a more \emph{specialized} ranking
policy. The teacher remains strong on several subsets, but becomes brittle
on a few domains that are farther from the effective support of the
instruction-following training distribution. In particular, the extremely
large drop on LitSearch indicates that the teacher's OOD weakness is
concentrated rather than uniform.

The final A3 student does not uniformly dominate the teacher either.
Instead, its main gain comes from \emph{repairing the teacher's largest
OOD failures}. A3 recovers dramatically on LitSearch, Touche, and
SciFact, while remaining slightly below the teacher on a number of
subsets such as ArguAna, DD\_2016, SciDocs, and Trec-Covid. This is
consistent with the paper's main interpretation: on-policy student
distillation does not simply inherit the Stage~1 teacher, but acts as a
corrective mechanism by evaluating supervision on rankings sampled from
the student's own policy.

\begin{table*}[t]
\centering
\small
\setlength{\tabcolsep}{6pt}
\begin{tabular}{lccccc}
\toprule
\textbf{MAIR-11 subset} & \textbf{Base-1B} & \textbf{Teacher GRPO} & \textbf{A3} & \textbf{Teacher--Base} & \textbf{A3--Teacher} \\
\midrule
ArguAna    & 0.374 & 0.600 & 0.462 & +0.227 & -0.138 \\
Core\_2017 & 0.656 & 0.629 & 0.675 & -0.027 & +0.046 \\
DD\_2016   & 0.530 & 0.701 & 0.527 & +0.171 & -0.175 \\
FiQA       & 0.832 & 0.931 & 0.965 & +0.099 & +0.034 \\
LitSearch  & 0.982 & 0.125 & 0.995 & -0.858 & +0.870 \\
NFCorpus   & 0.635 & 0.669 & 0.649 & +0.034 & -0.021 \\
Quora      & 0.951 & 1.000 & 0.978 & +0.049 & -0.022 \\
SciDocs    & 0.420 & 0.651 & 0.473 & +0.231 & -0.178 \\
SciFact    & 0.705 & 0.700 & 0.835 & -0.004 & +0.134 \\
Trec-Covid & 0.870 & 0.915 & 0.897 & +0.045 & -0.018 \\
Touche     & 0.930 & 0.751 & 0.941 & -0.179 & +0.190 \\
\bottomrule
\end{tabular}
\caption{Subset-level MAIR-11 diagnosis of Base-1B, Teacher GRPO, and A3
using nDCG@6. Teacher GRPO is stronger than Base-1B on most subsets, but
suffers a few large OOD failures, especially on LitSearch and Touche,
which dominate its aggregate MAIR-11 drop. A3 does not uniformly beat the
teacher; instead, its overall gain comes primarily from correcting those
largest OOD failures. Entries are within-subset means; aggregate
MAIR-11 values quoted in the text are query-micro averages over all
869 queries.}
\label{tab:appendix_teacher_base_a3_mair}
\end{table*}

\section{Cross-Architecture Generalization}
\label{sec:cross_architecture}

To test whether the Stage~2 training mechanism depends on the primary
Llama-Nemotron student backbone, we keep the strengthened ZeRank-2
teacher fixed and apply the same Stage~2 procedure to three alternative
student families: BGE-Reranker-v2-Gemma, MXBAI-Rerank-Large-v1,
and RankZephyr-7B. These models differ substantially in backbone scale
and reranking architecture, providing a direct test of student-side
architectural generality.

\paragraph{Architecture-specific student utilities.}
The teacher side is unchanged across all cross-architecture experiments:
the fixed Stage~1 ZeRank-2 teacher assigns each candidate the scaled
final-position \texttt{Yes}-token logit defined in
Stage~2 method, which is then used to construct the rank-normalized
relevances for the nDCG@6 reward and the teacher distribution for
distillation. Only the student-side mapping from the model's native
output to a scalar utility changes with architecture.

For BGE-Reranker-v2-Gemma, we use the raw \texttt{Yes}-token logit at
the final prompt position as the candidate utility, without applying a
sigmoid. For MXBAI-Rerank-Large-v1, we use the sequence-classification
relevance logit produced by the cross-encoder directly, again without
a sigmoid. RankZephyr-7B requires a different adaptation because it is
a listwise generative reranker rather than a pointwise scoring model.
We first greedily generate its ranking permutation, then teacher-force
the generated permutation and assign each document a utility equal to
the summed log-probability of its document-identifier token(s) at its
generated position. These document-level utilities provide a scalar
interface to the same training objective. In every architecture, the
resulting student utilities parameterize the Plackett--Luce policy used
for on-policy ranking samples and the student distribution used in the
KL distillation term. Thus, the Stage~2 objective and teacher reward
construction remain fixed; only the architecture-specific extraction of
student utilities changes.

\subsection{Validation Results}

Table~\ref{tab:cross_arch_val} reports validation performance before
and after applying Stage~2 training to each alternative student
architecture. All three backbones improve in both nDCG@6 and MRR@6,
showing that the gains are not specific to the primary
Llama-Nemotron student.

\begin{table*}[t]
\centering
\footnotesize
\setlength{\tabcolsep}{3.5pt}
\begin{tabular}{llccc}
\toprule
\textbf{Model (checkpoint)} & \textbf{Metric} &
\textbf{Before} & \textbf{After} & \textbf{$\Delta$} \\
\midrule
\multirow{2}{*}{BGE-Reranker-v2-Gemma (Step 300)}
& nDCG@6 & 0.7015 [0.6943, 0.7084] & 0.7335 [0.7261, 0.7407] & +0.0319 (+4.6\%) \\
& MRR@6  & 0.6805 [0.6727, 0.6884] & 0.7088 [0.7010, 0.7165] & +0.0283 (+4.2\%) \\
\midrule
\multirow{2}{*}{MXBAI-Rerank-Large-v1 (Step 600)}
& nDCG@6 & 0.7016 [0.6941, 0.7089] & 0.7247 [0.7174, 0.7320] & +0.0231 (+3.2\%) \\
& MRR@6  & 0.6753 [0.6675, 0.6834] & 0.6917 [0.6840, 0.6993] & +0.0163 (+2.4\%) \\
\midrule
\multirow{2}{*}{RankZephyr-7B (Step 200)}
& nDCG@6 & 0.7132 [0.7058, 0.7202] & 0.7416 [0.7322, 0.7482] & +0.0284 (+4.0\%) \\
& MRR@6  & 0.7884 [0.7672, 0.8042] & 0.8146 [0.8078, 0.8294] & +0.0262 (+3.3\%) \\
\bottomrule
\end{tabular}
\caption{Validation performance before and after Stage~2 training for
three alternative student architectures. The validation set contains
9,861 queries. Values are means with 95\% bootstrap confidence
intervals.}
\label{tab:cross_arch_val}
\end{table*}

\subsection{MAIR-11 Results}

Table~\ref{tab:cross_arch_mair} reports the corresponding MAIR-11
results. The improvements persist under distribution shift for all
three student architectures, with particularly large gains for
BGE-Reranker-v2-Gemma.

\begin{table*}[t]
\centering
\footnotesize
\setlength{\tabcolsep}{3.5pt}
\begin{tabular}{llccc}
\toprule
\textbf{Model (checkpoint)} & \textbf{Metric} &
\textbf{Before} & \textbf{After} & \textbf{$\Delta$} \\
\midrule
\multirow{2}{*}{BGE-Reranker-v2-Gemma (Step 300)}
& nDCG@6 & 0.4866 [0.4626, 0.5116] & 0.8173 [0.7991, 0.8350] & +0.3307 (+68.0\%) \\
& MRR@6  & 0.5550 [0.5269, 0.5827] & 0.8902 [0.8717, 0.9079] & +0.3352 (+60.4\%) \\
\midrule
\multirow{2}{*}{MXBAI-Rerank-Large-v1 (Step 600)}
& nDCG@6 & 0.6632 [0.6405, 0.6859] & 0.7017 [0.6791, 0.7239] & +0.0385 (+5.8\%) \\
& MRR@6  & 0.7194 [0.6935, 0.7443] & 0.7631 [0.7388, 0.7875] & +0.0437 (+6.1\%) \\
\midrule
\multirow{2}{*}{RankZephyr-7B (Step 200)}
& nDCG@6 & 0.7024 [0.6958, 0.7088] & 0.7382 [0.7336, 0.7438] & +0.0358 (+5.1\%) \\
& MRR@6  & 0.7612 [0.7494, 0.7648] & 0.8028 [0.7778, 0.8206] & +0.0416 (+5.5\%) \\
\bottomrule
\end{tabular}
\caption{MAIR-11 performance before and after Stage~2 training for
three alternative student architectures. MAIR-11 contains 869 queries
across 11 subsets.}
\label{tab:cross_arch_mair}
\end{table*}

All cross-architecture experiments use the same strengthened ZeRank-2
teacher produced by Stage~1. Repeating Stage~1 for multiple teacher
families would require additional costly multi-GPU teacher-training
runs and was outside our available compute budget. We use ZeRank-2
because it provides a strong open-weight instruction-following
reranker teacher and keeps these experiments aligned with the central
methodological question: whether reward-based on-policy distillation
can transfer a strengthened reranking policy across heterogeneous
student architectures. These experiments therefore establish
student-side architectural generality; teacher-side architectural
generality remains open.

\section{Full Set~B Numeric Ablation Results}
\label{sec:appendix_setb}

\begin{figure*}[!t]
\centering
\includegraphics[width=0.92\textwidth]{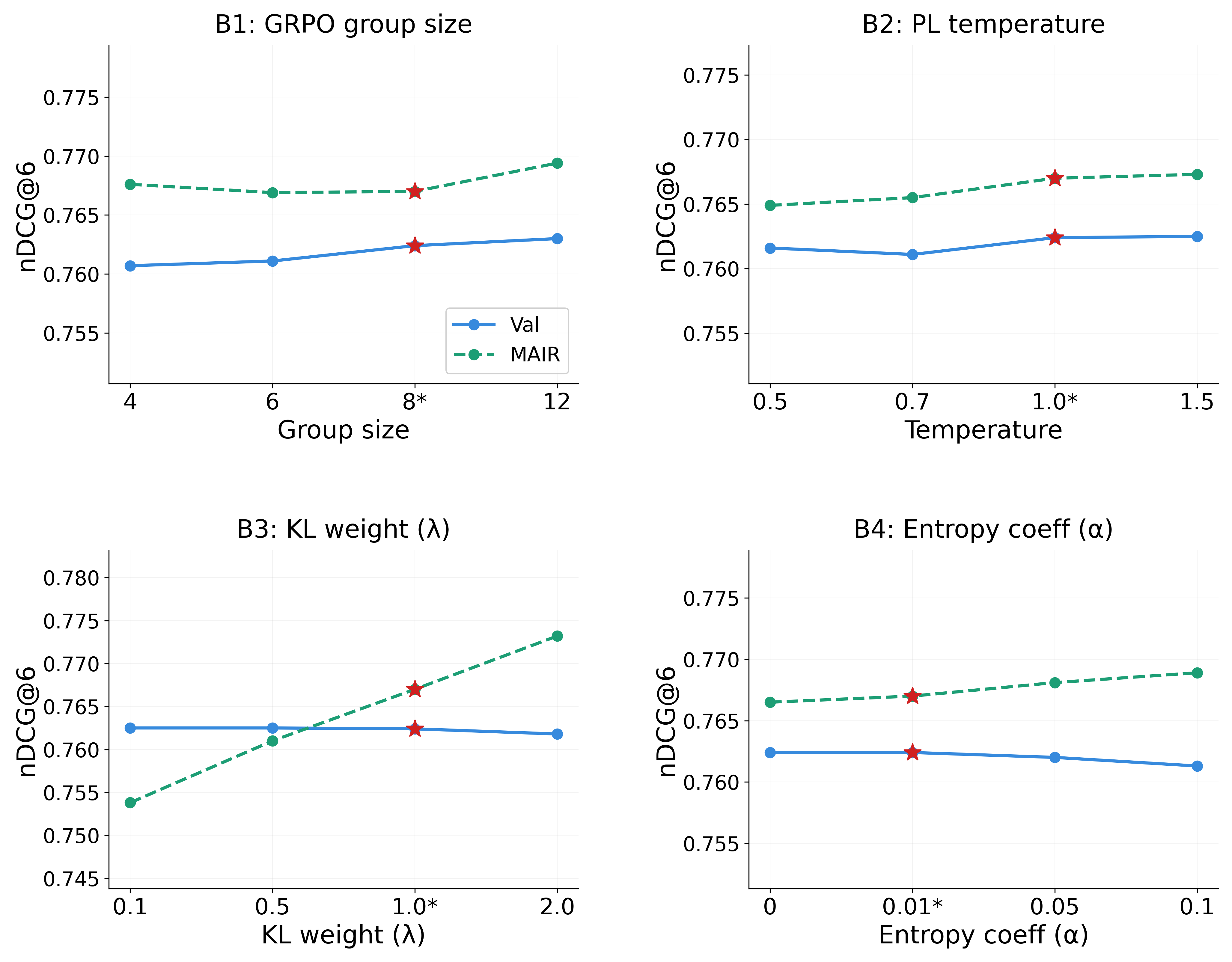}
\caption{Hyperparameter sensitivity of the default training setup.
Validation performance is nearly flat across all sweeps, while MAIR-11 shows modest sensitivity primarily to the KL weight. Red stars mark the default configuration used in the main results.}
\label{fig:setB_sensitivity}
\end{figure*}

The main paper summarizes Set~B qualitatively through the robustness
discussion. Here we provide the full plot
(Figure~\ref{fig:setB_sensitivity}) and numeric tables for
validation and MAIR-11 so that the stability claims can be inspected
directly. These results confirm that the method is not highly sensitive
to the exact default configuration: validation performance remains
nearly flat across all sweeps, while MAIR-11 shows only modest variation,
most visibly in the KL sweep.

\subsection{Validation Results}

Table~\ref{tab:appendix_setb_val} confirms that the in-distribution
benchmark is highly stable with respect to group size, sampling
temperature, KL weight, and entropy coefficient. The narrow spread of
scores suggests that the main gains are not coming from brittle tuning
around a single configuration.

\begin{table*}[!t]
\centering
\small
\setlength{\tabcolsep}{7pt}
\begin{tabular}{lccccc}
\toprule
\textbf{Variant} & \textbf{nDCG@6} & \textbf{SD} & \textbf{95\% CI} & \textbf{MRR@6} & \textbf{SD} \\
\midrule
B1: group\_size=4 & 0.7607 & 0.3729 & [0.753, 0.768] & 0.7454 & 0.3887 \\
B1: group\_size=6 & 0.7611 & 0.3730 & [0.754, 0.769] & 0.7458 & 0.3887 \\
B1: group\_size=12 & 0.7630 & 0.3731 & [0.756, 0.771] & 0.7484 & 0.3886 \\
B2: temp=0.5 & 0.7616 & 0.3729 & [0.754, 0.769] & 0.7465 & 0.3886 \\
B2: temp=0.7 & 0.7611 & 0.3729 & [0.754, 0.769] & 0.7459 & 0.3885 \\
B2: temp=1.5 & 0.7625 & 0.3730 & [0.755, 0.770] & 0.7478 & 0.3886 \\
B3: kl=0.1 & 0.7625 & 0.3731 & [0.755, 0.770] & 0.7476 & 0.3885 \\
B3: kl=0.5 & 0.7625 & 0.3731 & [0.755, 0.770] & 0.7476 & 0.3886 \\
B3: kl=2.0 & 0.7618 & 0.3731 & [0.755, 0.769] & 0.7473 & 0.3886 \\
B4: ent=0.0 & 0.7624 & 0.3731 & [0.755, 0.770] & 0.7476 & 0.3885 \\
B4: ent=0.05 & 0.7620 & 0.3730 & [0.755, 0.770] & 0.7470 & 0.3886 \\
B4: ent=0.1 & 0.7613 & 0.3730 & [0.754, 0.769] & 0.7460 & 0.3886 \\
\bottomrule
\end{tabular}
\caption{Full Set~B validation results. In-distribution performance
remains highly stable across all sweeps.}
\label{tab:appendix_setb_val}
\end{table*}

\subsection{MAIR-11 Results}

Table~\ref{tab:appendix_setb_mair} shows the same sweeps on MAIR-11. While the overall method remains stable, MAIR-11 displays somewhat stronger sensitivity to the KL coefficient than the validation set. In particular, larger KL values improve OOD performance, consistent with our interpretation in the main text that KL is not the main source of learning but can still improve generalization stability.

\begin{table*}[!t]
\centering
\small
\setlength{\tabcolsep}{7pt}
\begin{tabular}{lccccc}
\toprule
\textbf{Variant} & \textbf{nDCG@6} & \textbf{SD} & \textbf{95\% CI} & \textbf{MRR@6} & \textbf{SD} \\
\midrule
B1: group\_size=4 & 0.7676 & 0.3275 & [0.745, 0.789] & 0.8319 & 0.3318 \\
B1: group\_size=6 & 0.7669 & 0.3279 & [0.745, 0.789] & 0.8305 & 0.3330 \\
B1: group\_size=12 & 0.7694 & 0.3261 & [0.748, 0.791] & 0.8328 & 0.3309 \\
B2: temp=0.5 & 0.7649 & 0.3267 & [0.743, 0.787] & 0.8306 & 0.3313 \\
B2: temp=0.7 & 0.7655 & 0.3268 & [0.744, 0.788] & 0.8305 & 0.3315 \\
B2: temp=1.5 & 0.7673 & 0.3283 & [0.745, 0.789] & 0.8314 & 0.3327 \\
B3: kl=0.1 & 0.7538 & 0.3360 & [0.732, 0.776] & 0.8142 & 0.3418 \\
B3: kl=0.5 & 0.7610 & 0.3297 & [0.739, 0.783] & 0.8243 & 0.3365 \\
B3: kl=2.0 & 0.7732 & 0.3210 & [0.752, 0.795] & 0.8378 & 0.3258 \\
B4: ent=0.0 & 0.7665 & 0.3285 & [0.745, 0.788] & 0.8301 & 0.3328 \\
B4: ent=0.05 & 0.7681 & 0.3262 & [0.746, 0.790] & 0.8316 & 0.3311 \\
B4: ent=0.1 & 0.7689 & 0.3251 & [0.747, 0.791] & 0.8326 & 0.3298 \\
\bottomrule
\end{tabular}
\caption{Full Set~B MAIR-11 results. OOD performance is broadly stable,
with the most visible sensitivity appearing in the KL sweep.}
\label{tab:appendix_setb_mair}
\end{table*}

\section{Per-Dataset Validation Breakdown Across All Baselines}
\label{sec:appendix_perdataset}

Aggregate validation results in the main paper hide meaningful
differences across datasets. Table~\ref{tab:appendix_perdataset_full}
provides the per-dataset validation breakdown across the major
baselines and the final distilled model. For readability, the models
are divided into two panels, with A3 repeated in both panels to provide
a common reference.

This view makes two points clear. First, the distilled student is not
winning only because of a single dataset; it remains competitive across
web search, code, mathematics, instruction-following, and multi-hop
retrieval settings. Second, different baselines exhibit complementary
strengths. For example, Jina Reranker v2 is particularly strong on
MS MARCO and Robust04, Cohere Rerank v4.0-fast is strongest on
MetaMath and LeetCode, and the 7B RL-trained rerankers reach saturation
on FollowIR and InstructIR.

Rank-R1-7B and REARANK-7B are evaluated using their native
sliding-window listwise inference procedure (window size 20, step size
10), while the remaining models use their corresponding primary
evaluation pipelines. All systems operate on the same frozen validation
candidate pools. The overall row reports the query-micro average across
all 9,861 validation queries.

\begin{table*}[!t]
\centering
\small
\setlength{\tabcolsep}{7pt}

\textbf{Panel A: Internal and open-weight reference models}

\vspace{3pt}

\begin{tabular}{lcccccc}
\toprule
\textbf{Dataset} &
\textbf{Qwen3-4B} &
\textbf{Base-1B} &
\textbf{ZeRank-2} &
\textbf{BGE v2} &
\textbf{Teacher GRPO} &
\textbf{A3 (ours)} \\
\midrule

FollowIR
& 0.910
& 0.947
& \textbf{1.000}
& 0.962
& \textbf{1.000}
& 0.977 \\

InstructIR
& 0.760
& 0.956
& \textbf{1.000}
& \textbf{1.000}
& \textbf{1.000}
& 0.999 \\

InfoSearch
& 0.829
& 0.835
& 0.831
& 0.831
& 0.843
& 0.831 \\

MS MARCO
& 0.916
& 0.811
& 0.862
& 0.847
& 0.897
& 0.919 \\

MetaMath
& 0.728
& 0.797
& 0.906
& 0.950
& 0.926
& 0.953 \\

LeetCode
& 0.717
& 0.745
& 0.712
& 0.733
& 0.724
& 0.789 \\

Robust04
& 0.952
& 0.605
& 0.598
& 0.578
& 0.620
& 0.662 \\

WebQA
& 0.255
& 0.444
& 0.422
& \textbf{0.458}
& 0.435
& \textbf{0.458} \\

\midrule
\textbf{Overall (micro)}
& 0.656
& 0.697
& 0.722
& 0.731
& 0.742
& \textbf{0.762} \\

\bottomrule
\end{tabular}

\vspace{8pt}

\textbf{Panel B: External baselines and 7B RL-trained rerankers}

\vspace{3pt}

\begin{tabular}{lcccccc}
\toprule
\textbf{Dataset} &
\textbf{Cohere 3.5} &
\textbf{Cohere 4-fast} &
\textbf{Jina v2} &
\textbf{Rank-R1-7B} &
\textbf{REARANK-7B} &
\textbf{A3 (ours)} \\
\midrule

FollowIR
& 0.970
& \textbf{1.000}
& 0.993
& \textbf{1.000}
& \textbf{1.000}
& 0.977 \\

InstructIR
& \textbf{1.000}
& \textbf{1.000}
& \textbf{1.000}
& \textbf{1.000}
& \textbf{1.000}
& 0.999 \\

InfoSearch
& \textbf{0.848}
& 0.844
& 0.843
& 0.813
& 0.808
& 0.831 \\

MS MARCO
& 0.879
& 0.874
& \textbf{0.943}
& 0.902
& 0.891
& 0.919 \\

MetaMath
& 0.948
& \textbf{0.971}
& 0.822
& 0.922
& 0.929
& 0.953 \\

LeetCode
& 0.778
& \textbf{0.831}
& 0.754
& 0.759
& 0.762
& 0.789 \\

Robust04
& 0.660
& 0.819
& \textbf{0.970}
& 0.768
& 0.774
& 0.662 \\

WebQA
& 0.455
& 0.453
& 0.435
& 0.446
& 0.452
& \textbf{0.458} \\

\midrule
\textbf{Overall (micro)}
& 0.746
& 0.749
& 0.761
& 0.750
& 0.748
& \textbf{0.762} \\

\bottomrule
\end{tabular}

\caption{Per-dataset validation nDCG@6 across major baselines and the
final distilled model. Models are divided into two panels for
readability, with A3 repeated in both panels as a common reference.
Rank-R1-7B and REARANK-7B use their native sliding-window listwise
inference procedure (window size 20, step size 10), whereas the
remaining models use their corresponding primary evaluation pipelines.
All systems operate on the same frozen validation candidate pools.
\textbf{Bold} indicates the highest point estimate in each row across
both panels, with ties highlighted. Overall (micro) denotes the
query-micro average across all 9,861 validation queries.}
\label{tab:appendix_perdataset_full}
\end{table*}

The per-dataset results show that the distilled student does not derive
its aggregate performance from a single validation source. A3 remains
competitive across all eight datasets and obtains the highest or tied
highest point estimate on WebQA, while remaining particularly strong on
MS MARCO, MetaMath, and LeetCode. At the same time, other models retain
clear dataset-specific strengths: Jina Reranker v2 is strongest on
MS MARCO and Robust04, Cohere Rerank v4.0-fast is strongest on MetaMath
and LeetCode, and several models reach saturation on FollowIR and
InstructIR.

The comparison with the released 7B RL-trained rerankers shows a
similarly heterogeneous pattern. A3 exceeds both Rank-R1-7B and
REARANK-7B on InfoSearch, MS MARCO, MetaMath, LeetCode, and WebQA,
whereas both 7B rerankers are stronger on Robust04 and reach 1.000 on
FollowIR and InstructIR. Under query-micro aggregation across the full
9,861-query validation benchmark, A3 reaches 0.762 nDCG@6, compared
with 0.750 for Rank-R1-7B and 0.748 for REARANK-7B, while using a
substantially smaller model. These results support the interpretation
that the distilled model provides a strong overall
quality--robustness tradeoff rather than uniformly dominating every
individual validation subset.

\subsection{Per-Subset Comparison of A3, A4, and A5}
\label{sec:appendix_a345}

To better understand the role of the KL and entropy terms, we compare
A3 (full objective), A4 (without KL), and A5 (without entropy) at the
per-subset level on MAIR-11 (Table~\ref{tab:appendix_a345_mair}). The three variants are very close in
aggregate, and the subset-level picture remains mixed: wins are split
across domains, and none of the three variants consistently dominates
the others. This supports the interpretation in the main paper that
policy-gradient training with teacher-derived soft rewards is the
primary source of improvement, while KL and entropy act as auxiliary
stabilizers rather than indispensable components.

A4 is slightly strongest overall on MAIR-11, while A3 and A5 remain very
close to one another. A3 remains strongest on several OOD subsets,
including Core\_2017, DD\_2016, Trec-Covid, and LitSearch, while A4 is
strongest on others, including ArguAna, FiQA, Quora, and Touche. A5 is
also competitive across most subsets. Overall, the absence of a
consistent winner suggests that the differences among A3, A4, and A5
are modest relative to the larger gap between on-policy student-driven
distillation and the offline or off-policy alternatives.

\begin{table*}[!t]
\centering
\small
\setlength{\tabcolsep}{6pt}
\begin{tabular}{lcccccc}
\toprule
\multirow{2}{*}{\textbf{MAIR-11 subset}} &
\multicolumn{3}{c}{\textbf{nDCG@6}} &
\multicolumn{3}{c}{\textbf{MRR@6}} \\
\cmidrule(lr){2-4} \cmidrule(lr){5-7}
& \textbf{A3} & \textbf{A4} & \textbf{A5}
& \textbf{A3} & \textbf{A4} & \textbf{A5} \\
\midrule
ArguAna    & 0.4625 & \textbf{0.5444} & 0.4670 & 0.4147 & \textbf{0.4975} & 0.4178 \\
Core\_2017 & \textbf{0.6745} & 0.6720 & 0.6738 & \textbf{1.0000} & \textbf{1.0000} & \textbf{1.0000} \\
DD\_2016   & \textbf{0.5267} & 0.5171 & 0.5147 & \textbf{0.8450} & 0.8317 & 0.8367 \\
FiQA       & 0.9646 & \textbf{0.9687} & 0.9645 & 0.9833 & \textbf{0.9867} & 0.9833 \\
LitSearch  & \textbf{0.9950} & 0.9943 & 0.9943 & \textbf{0.9933} & 0.9900 & 0.9925 \\
NFCorpus   & 0.6487 & 0.6334 & \textbf{0.6505} & 0.7522 & 0.7278 & \textbf{0.7588} \\
Quora      & 0.9784 & \textbf{0.9788} & 0.9788 & 0.9728 & \textbf{0.9733} & \textbf{0.9733} \\
SciDocs    & 0.4729 & 0.4355 & \textbf{0.4755} & 0.6640 & 0.6233 & \textbf{0.6717} \\
SciFact    & \textbf{0.8346} & 0.8239 & 0.8303 & \textbf{0.8103} & 0.8033 & 0.8078 \\
Trec-Covid & \textbf{0.8970} & 0.8872 & 0.8922 & \textbf{1.0000} & \textbf{1.0000} & \textbf{1.0000} \\
Touche     & 0.9415 & \textbf{0.9493} & 0.9431 & \textbf{1.0000} & \textbf{1.0000} & \textbf{1.0000} \\
\midrule
\textbf{Overall MAIR-11} & 0.7670 & \textbf{0.7688} & 0.7665 & 0.8289 & 0.8282 & \textbf{0.8301} \\
\bottomrule
\end{tabular}
\caption{Per-subset MAIR-11 comparison for A3 (full objective), A4 (no
KL), and A5 (no entropy), reported for both nDCG@6 and MRR@6. All three
variants are very close in aggregate, and wins remain split across
subsets rather than dominated by a single configuration. Per-subset
entries are within-subset means; the Overall MAIR-11 row is the
query-micro average over all 869 queries and therefore does not equal
the unweighted mean of the subset entries.}
\label{tab:appendix_a345_mair}
\end{table*}

\subsection{Paired Statistical Tests for Key Variant Comparisons}
\label{sec:appendix_wilcoxon}

To complement the aggregate bootstrap confidence intervals reported in
the main paper, we additionally compare key variants using paired
Wilcoxon signed-rank tests over query-level nDCG@6 scores. This
analysis focuses on the two most important comparisons for the paper's
claims: A3 versus A1, which tests whether on-policy distillation
improves over offline KD, and A3 versus A4, which tests whether the KL
term materially changes the behavior of the full objective.

Table~\ref{tab:appendix_wilcoxon} shows two qualitatively different
outcomes. First, A3 shows a small positive paired difference relative
to A1 on validation and a substantially larger, statistically
significant advantage on MAIR-11. On validation, the Wilcoxon
$p$-value is extreme ($<10^{-320}$, below float64 resolution)
even though the bootstrap CI on the mean difference includes
zero; this is not a contradiction, as the signed-rank test
measures the consistency of the sign of per-query differences
across 9{,}861 paired queries, whereas the CI reflects the
magnitude of the mean effect. A highly consistent but very
small per-query advantage therefore yields an extreme $p$
alongside a near-zero mean difference, and we accordingly
read the validation comparison as directionally reliable but
practically negligible, with the substantive A3--A1 gap
emerging under distribution shift. This pattern is consistent with the
main result that the benefit of on-policy student-driven distillation
becomes most pronounced under distribution shift. Second, A4 is
slightly but significantly stronger than A3 in paired tests, especially
on validation. At the same time, the per-subset MAIR-11 results do not
show uniform dominance: A3 is strongest on several subsets, while A4 is
strongest on others. We therefore interpret A3 as a useful reference
formulation of the full objective rather than as the uniquely
best-performing member of this closely related family.

\begin{table*}[t]
\centering
\small
\setlength{\tabcolsep}{8pt}
\begin{tabular}{llrrrr}
\toprule
\textbf{Comparison} & \textbf{Split} & \textbf{Mean diff.} & \textbf{95\% CI} & \textbf{Wilcoxon $p$ (2-sided)} & \textbf{Direction} \\
\midrule
A3 vs A1 & Validation & +0.0016 & [-0.0014, 0.0046] & $<10^{-320}$ & A3 $>$ A1 \\ \\
A3 vs A1 & MAIR-11 & +0.0220 & [0.0151, 0.0291] & $1.95\times10^{-9}$ & A3 $>$ A1 \\
A3 vs A4 & Validation & -0.0064 & [-0.0078, -0.0050] & $5.91\times10^{-33}$ & A3 $<$ A4 \\
A3 vs A4 & MAIR-11 & -0.0043 & [-0.0076, -0.0012] & $2.69\times10^{-2}$ & A3 $<$ A4 \\
\bottomrule
\end{tabular}
\caption{Paired Wilcoxon signed-rank tests over query-level nDCG@6
scores for the two most important variant comparisons. Positive mean
differences favor A3; negative mean differences favor the comparison
variant.}
\label{tab:appendix_wilcoxon}
\end{table*}

\begin{figure*}[t]
\centering
\includegraphics[width=0.88\textwidth]{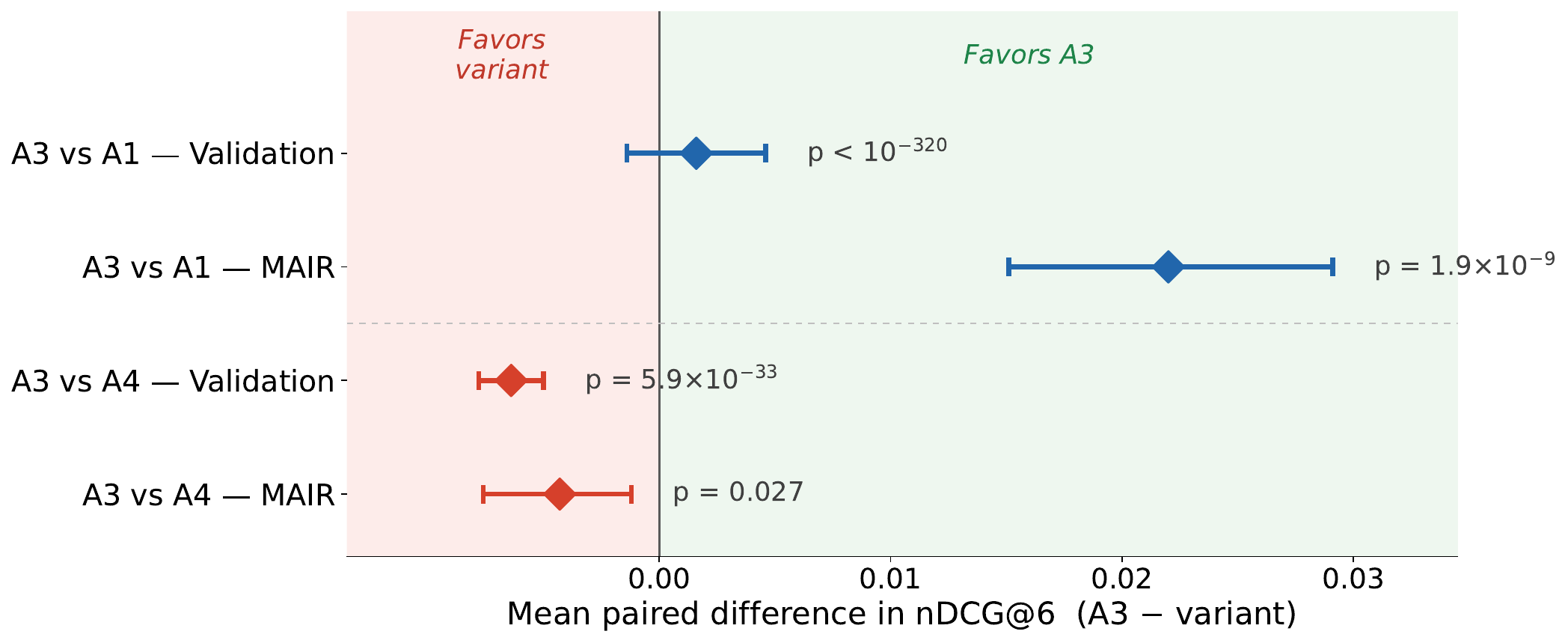}
\caption{Forest plot of paired query-level nDCG@6 effect estimates for
the two key variant comparisons. Positive values favor A3; negative
values favor the comparison variant. The A3--A1 effect is small on
validation and substantially larger on MAIR-11, while A4 is slightly
stronger than A3 in the paired comparisons. Horizontal bars show
bootstrap 95\% confidence intervals.}
\label{fig:appendix_wilcoxon_forest}
\end{figure*}

Figure~\ref{fig:appendix_wilcoxon_forest} provides a compact visual
summary of these paired effect estimates. The A3--A1 difference is
small on validation but substantially larger on MAIR-11, whereas the
A3--A4 comparison favors A4 by a relatively small margin on both
splits.

\subsection{Multi-seed stability of A3, A4, and A5}
\label{sec:appendix_multiseed}

A possible concern in Section~\ref{sec:ablation} is that the close ordering
among A3, A4, and A5 could reflect single-seed noise rather than a stable
relationship between these objective variants. We therefore reran the
tightly clustered A3/A4/A5 family with 3 random seeds and evaluated each
run on both the validation benchmark and MAIR-11.

We focus on A3, A4, and A5 because these are the variants whose
single-seed results are closest in both validation and OOD evaluation,
and hence are the only ones for which training-seed variance could
materially affect the interpretation. In contrast, the larger gaps for
A1, A2, A6, and A8 are already well separated from A3 in the main
results and are less likely to be explained by small seed-level
fluctuations.

Absolute values in Table~\ref{tab:multiseed_a3_a4_a5} are not
directly comparable to Tables~\ref{tab:main_results}
and~\ref{tab:mair_compact}: the multi-seed reruns select the
best-validation checkpoint per run, rather than the fixed final
checkpoint reported in the main tables. This table is therefore
intended for comparing A3, A4, and A5 \emph{to one another} under a
shared protocol across seeds, not for cross-table comparison of
absolute scores.

\begin{figure*}[t]
    \centering
    \begin{subfigure}[t]{0.49\textwidth}
        \centering
        \includegraphics[width=\textwidth]{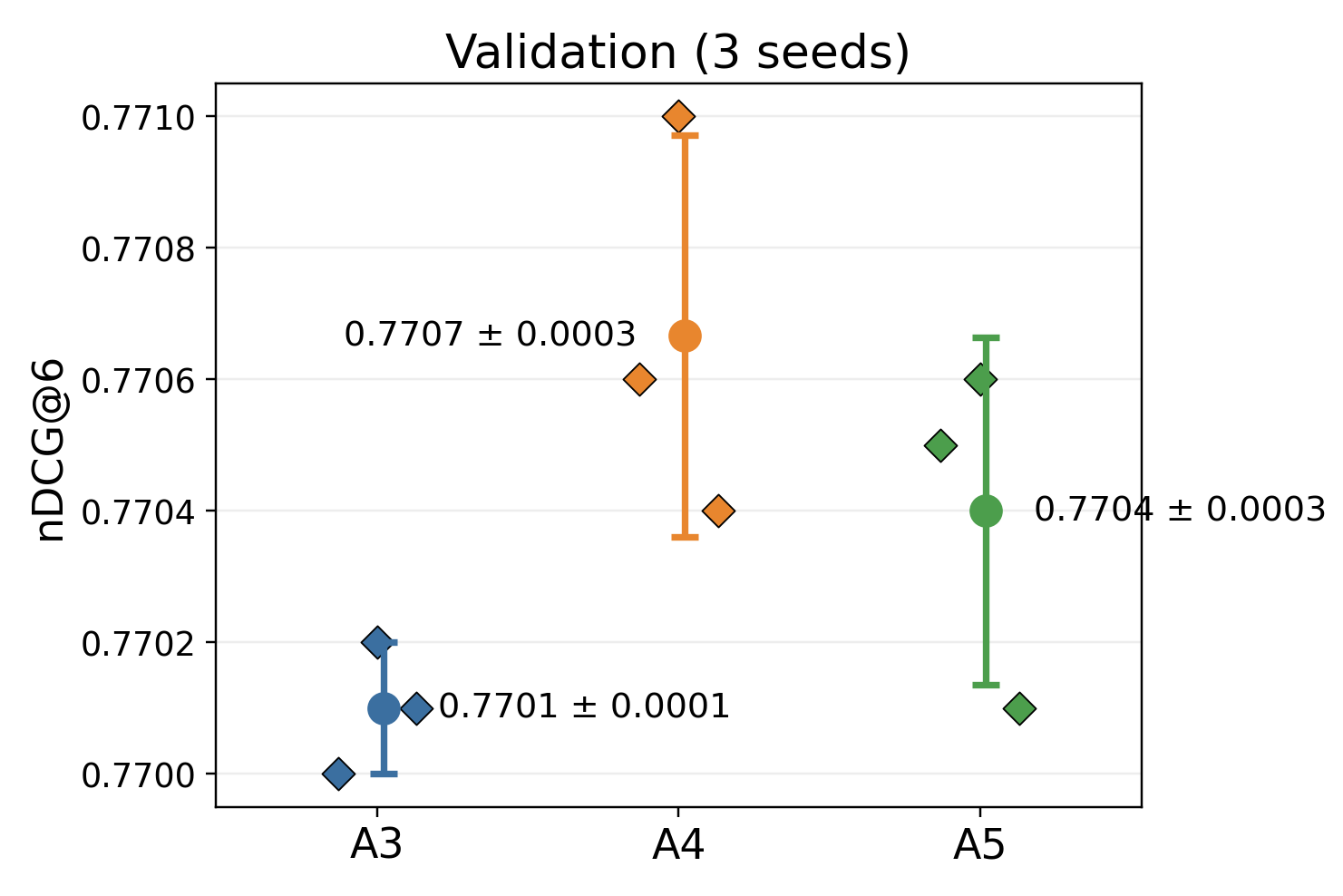}
        \caption{Validation benchmark over 3 random seeds.}
        \label{fig:multiseed_a3_a4_a5_val}
    \end{subfigure}
    \hfill
    \begin{subfigure}[t]{0.49\textwidth}
        \centering
        \includegraphics[width=\textwidth]{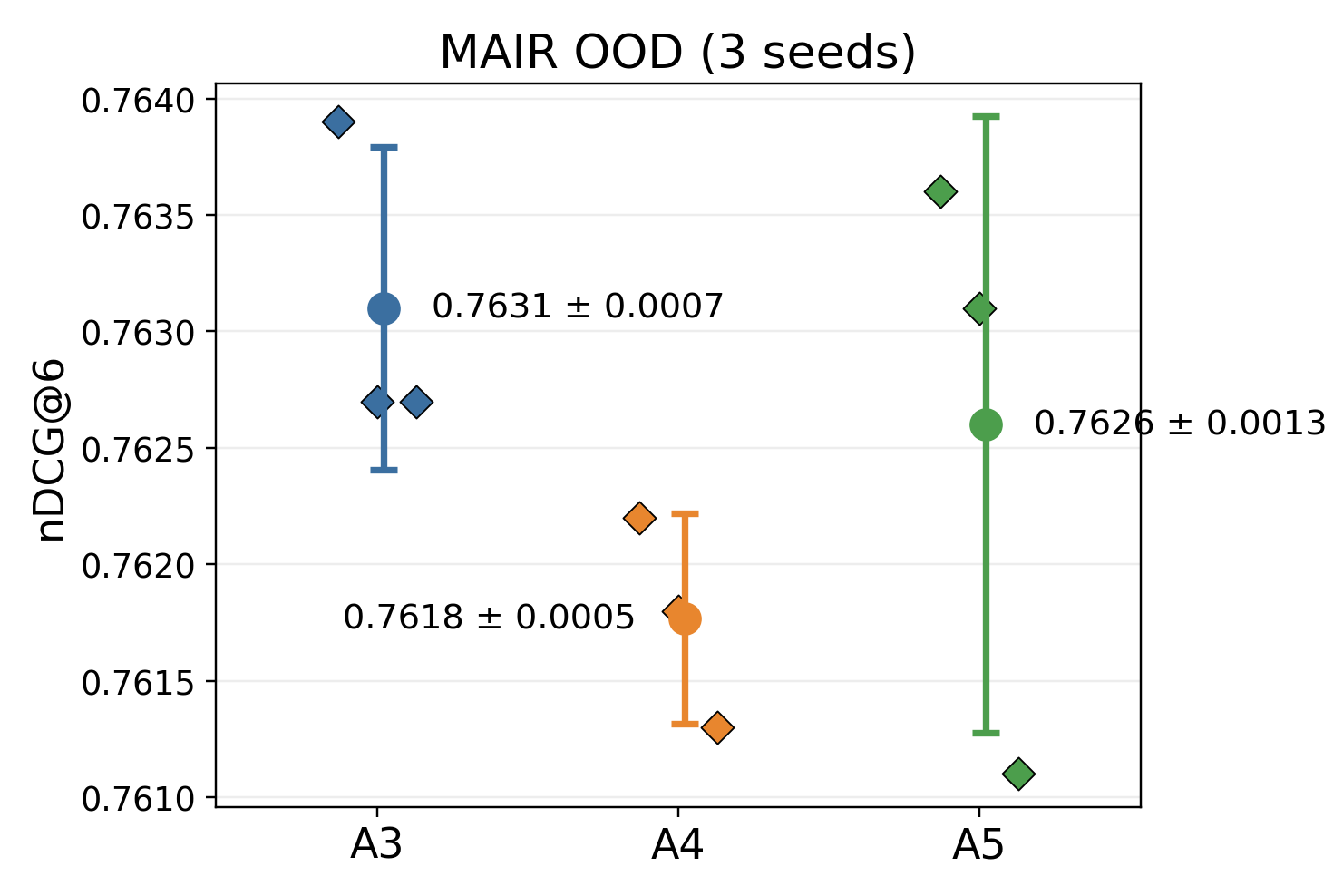}
        \caption{MAIR-11 OOD benchmark over 3 random seeds.}
        \label{fig:multiseed_a3_a4_a5_mair}
    \end{subfigure}
    \caption{Multi-seed stability of the tightly clustered A3/A4/A5 family over 3 random seeds.
    Dots show individual seeds; circles and error bars show mean $\pm$ standard deviation across seeds.
    A4 is marginally strongest on the validation benchmark, while A3 is marginally strongest on MAIR-11.
    The small spread across seeds supports our interpretation that these three variants are closely related,
    and that the main gain comes from student-driven on-policy learning with soft teacher rewards, while
    KL and entropy act as secondary stabilizers.}
    \label{fig:multiseed_a3_a4_a5}
\end{figure*}

Table~\ref{tab:multiseed_a3_a4_a5} reports the aggregate 3-seed results,
and Figure~\ref{fig:multiseed_a3_a4_a5} visualizes the same comparison
for validation and MAIR-11. The multi-seed results confirm that A3, A4,
and A5 remain tightly clustered across seeds. A4 is marginally strongest
on the validation benchmark, while A3 is marginally strongest on MAIR-11,
with A5 remaining close to both. This reinforces our interpretation that
the main gain comes from student-driven on-policy learning with soft
teacher rewards, while the KL and entropy terms act as secondary
stabilizers rather than determining the core improvement. We therefore
treat A3/A4/A5 as a closely related family of strong variants rather
than claiming a single uniformly dominant configuration.

\begin{table*}[t]
\centering
\small
\setlength{\tabcolsep}{10pt}
\begin{tabular}{lccccc}
\toprule
\textbf{Variant} & \textbf{\# Seeds} & \textbf{Val nDCG@6} & \textbf{Val MRR@6} & \textbf{MAIR-11 nDCG@6} & \textbf{MAIR-11 MRR@6} \\
\midrule
A3 & 3 & 0.7701 $\pm$ 0.0001 & 0.7449 $\pm$ 0.0002 & 0.7631 $\pm$ 0.0007 & 0.8252 $\pm$ 0.0006 \\
A4 & 3 & 0.7707 $\pm$ 0.0003 & 0.7490 $\pm$ 0.0004 & 0.7618 $\pm$ 0.0005 & 0.8229 $\pm$ 0.0011 \\
A5 & 3 & 0.7704 $\pm$ 0.0003 & 0.7486 $\pm$ 0.0004 & 0.7626 $\pm$ 0.0013 & 0.8247 $\pm$ 0.0022 \\
\bottomrule
\end{tabular}
\caption{Aggregate multi-seed results for the tightly clustered A3/A4/A5 variants.
Values are mean $\pm$ standard deviation across 3 random seeds, computed under
the multi-seed rerun protocol (best-validation checkpoint selection) and therefore not directly comparable to
Tables~\ref{tab:main_results} and~\ref{tab:mair_compact}; see text.}
\label{tab:multiseed_a3_a4_a5}
\end{table*}

\subsection{Qualitative Comparisons}
\label{sec:appendix_qualitative}

Table~\ref{tab:appendix_qualitative_examples} shows four representative
qualitative comparisons centered on the final A3 student. We include one
example each for offline KD (A1), the Stage~1 teacher, hard-label
training (A6), and off-policy student distillation (A2). In all four
cases, A3 flips the rank-1 decision: the relevant document is promoted
to the top position, whereas the comparison model assigns rank~1 to a
near-miss or distractor. These examples are intended as illustrations of
the aggregate trends in the main paper rather than as standalone
evidence.

Two patterns recur across the examples. First, many failures are not
gross topical errors but \emph{ranking} errors: the competing model
retrieves a plausible candidate set but places the best instruction- or
claim-aligned passage below less relevant distractors. Second, the A3
student often improves by making finer distinctions among highly similar
candidates rather than by discovering an entirely different document.
This is consistent with the broader empirical picture in the paper:
student-driven on-policy distillation improves the ordering of candidate
documents even when the relevant passage is already present in the pool.

\begin{table*}[t]
\centering
\small
\setlength{\tabcolsep}{4pt}
\begin{tabular}{p{0.13\textwidth} p{0.24\textwidth} p{0.24\textwidth} p{0.24\textwidth} p{0.11\textwidth}}
\toprule
\textbf{Comparison / Dataset} & \textbf{Query + Instruction} & \textbf{A3 top-3} & \textbf{Other model top-3} & \textbf{Takeaway} \\
\midrule

\textbf{A1 vs A3} \newline
\textit{InfIR\_robust04}
&
\textbf{Query:} Explore the history and evolution of the property market and its key figures. \newline
\textbf{Instruction:} Look for articles or blog posts discussing the rise of property developers, their influence, and the changes in the property market over time.
&
\textbf{1.} ``history of the property market. Developers became rich, influential and famous \ldots'' (\textbf{rel=1}) \newline
2. ``The rapid rise of property development in urban areas has led to a complex \ldots'' (rel=0) \newline
3. ``The history and evolution of the property market is a complex narrative \ldots'' (rel=0)
&
\textbf{1.} ``The rapid rise of property development in urban areas has led to a complex \ldots'' (rel=0) \newline
2. ``The history and evolution of the property market is a complex narrative \ldots'' (rel=0) \newline
3. ``history of the property market. Developers became rich, influential and famous \ldots'' (\textbf{rel=1})
&
A3 promotes the only clearly instruction-aligned passage to rank~1,
whereas offline KD under-ranks it behind broader but less targeted
property-market summaries. \\

\midrule

\textbf{Teacher vs A3} \newline
\textit{InfoSearch}
&
\textbf{Query:} Wildlife flourishing in uninhabited areas around Fukushima? I would value a blog entry discussing this matter. \newline
\textbf{Instruction:} Determine whether a document is relevant to the query.
&
\textbf{1.} ``Wildlife Thriving in Fukushima's Uninhabited Zones: A Decade After the Disaster \ldots'' (\textbf{rel=1}) \newline
2. ``Nearly a decade after Japan's Fukushima nuclear disaster, researchers have discovered \ldots'' (rel=0) \newline
3. ``User123: I was really surprised to hear that wildlife is flourishing \ldots'' (rel=0)
&
\textbf{1.} ``User123: I was really surprised to hear that wildlife is flourishing \ldots'' (rel=0) \newline
2. ``Nearly a decade after Japan's Fukushima nuclear disaster, researchers have discovered \ldots'' (rel=0) \newline
3. ``Wildlife Thriving in Fukushima's Uninhabited Zones: A Decade After the Disaster \ldots'' (\textbf{rel=1})
&
A3 correctly prioritizes the blog-style document that matches both topic
and form, while the teacher prefers weaker near-miss passages and buries
the relevant one at rank~3. \\

\midrule

\textbf{A6 vs A3} \newline
\textit{MAIR\_ArguAna}
&
\textbf{Query:} More women in the labour market leads to higher GDP \ldots \newline
\textbf{Instruction:} Given a claim, find documents that refute the claim.
&
\textbf{1.} ``gender house believes gender quotas EU are advantageous economies member states \ldots'' (\textbf{rel=1}) \newline
2. ``Most corporations, in almost every country on the earth will not even offer \ldots'' (rel=0) \newline
3. ``Slum dwellers in Nairobi are shown to pay high rents \ldots'' (rel=0)
&
\textbf{1.} ``Most corporations, in almost every country on the earth will not even offer \ldots'' (rel=0) \newline
2. ``Slum dwellers in Nairobi are shown to pay high rents \ldots'' (rel=0) \newline
3. ``gender house believes gender quotas EU are advantageous economies member states \ldots'' (\textbf{rel=1})
&
Under OOD argumentative retrieval, A3 moves the true counterargument to
rank~1, while the hard-label variant overweights semantically related but
non-refuting distractors. \\

\midrule

\textbf{A2 vs A3} \newline
\textit{InfIR\_msmarco}
&
\textbf{Query:} dame mas gasolina meaning \newline
\textbf{Instruction:} Relevant documents should explain the meaning of
the phrase, including its literal translation and contextual
interpretation.
&
\textbf{1.} ``Dame Mas Gasolina. It actually means gasoline but in this song it means that \ldots'' (\textbf{rel=1}) \newline
2. ``The phrase `dame mas gasolina' literally translates to `give me more gasoline' \ldots'' (rel=0) \newline
3. ``The phrase `dame más gasolina,' while literally meaning `give me more gasoline' \ldots'' (rel=0)
&
\textbf{1.} ``The phrase `dame más gasolina,' while literally meaning `give me more gasoline' \ldots'' (rel=0) \newline
2. ``The phrase `dame mas gasolina' literally translates to `give me more gasoline' \ldots'' (rel=0) \newline
3. ``Dame Mas Gasolina. It actually means gasoline but in this song it means that \ldots'' (\textbf{rel=1})
&
A3 places the best contextual explanation at rank~1, whereas the
off-policy variant prefers generic translation-like distractors and
under-ranks the most complete answer. \\

\bottomrule
\end{tabular}
\caption{Representative qualitative comparisons drawn from the
automatically mined example pool. Each case is a rank-1 flip where A3
promotes the relevant document above the competing model. Together, the
examples illustrate four distinct effects discussed in the paper:
improvement over offline KD, correction of teacher ranking errors,
greater robustness than hard-label training, and better ranking than the
off-policy student variant.}
\label{tab:appendix_qualitative_examples}
\end{table*}

\subsection{Offline KD vs On-Policy Distillation}
\label{sec:appendix_a1a3}

A1 and A3 are intentionally close on the in-distribution validation
benchmark: A1 reaches 0.7608 nDCG@6 and A3 reaches 0.7624, with heavily
overlapping confidence intervals. We therefore do not interpret the
benefit of on-policy distillation as a stronger in-distribution fitting
mechanism than offline KD. Instead, the distinction becomes visible
under distribution shift. On MAIR-11, A3 improves over A1 from 0.7212 to
0.7670 nDCG@6 and from 0.7860 to 0.8289 MRR@6, corresponding to gains
of 4.6 and 4.3 points, respectively. This result indicates that the
student benefits from being trained on rankings sampled from its own
policy rather than only matching a fixed teacher supervision
distribution.

This pattern is consistent with the central interpretation of the paper.
Offline KD transfers the teacher's preferences efficiently, but only on
the teacher-supported ranking distribution. In contrast, on-policy
distillation exposes the student to its \emph{own} ranking errors and
asks the teacher to evaluate those rankings. The result is not a large
in-distribution gain over offline KD, but a more robust student under
distribution shift. For this reason, we treat A3 not as a universally
better optimizer than A1, but as a more generalization-oriented
distillation strategy.

\section{LLM Judge Validation}
\label{sec:appendix_judge}

Because Stage~1 relies on judge-derived reward signals, it is important
to verify that the supervision is directionally consistent across strong
LLM evaluators. We therefore compare five judges (Gemini~2.5, Claude
Sonnet~4, GPT-4o, Mistral, and OSS-20B) on criteria directly relevant to
instruction-following reranking.

\begin{figure*}[t]
\centering
\begin{subfigure}[t]{0.49\textwidth}
    \centering
    \includegraphics[width=\textwidth]{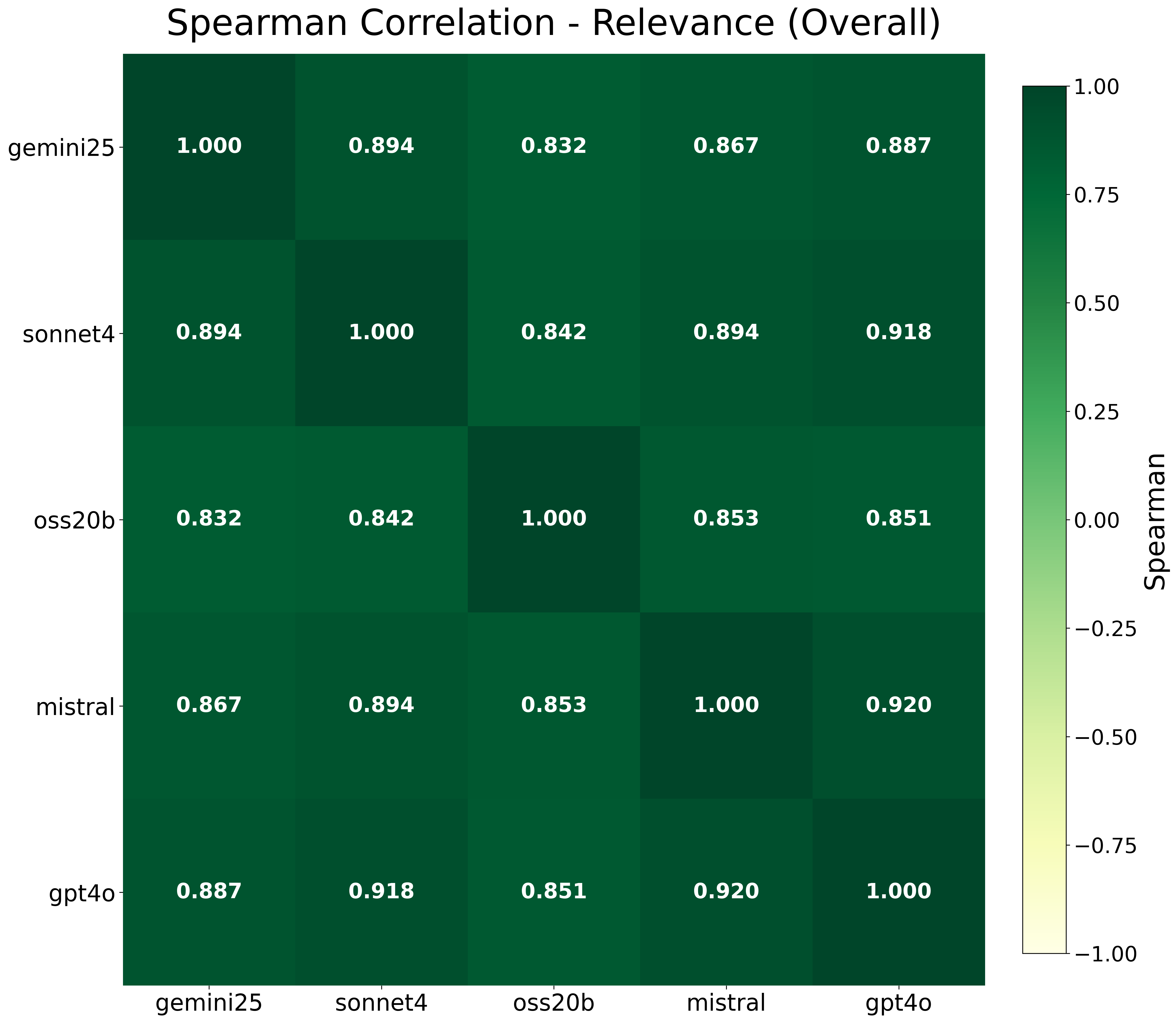}
    \caption{Spearman correlation: relevance}
\end{subfigure}
\hfill
\begin{subfigure}[t]{0.49\textwidth}
    \centering
    \includegraphics[width=\textwidth]{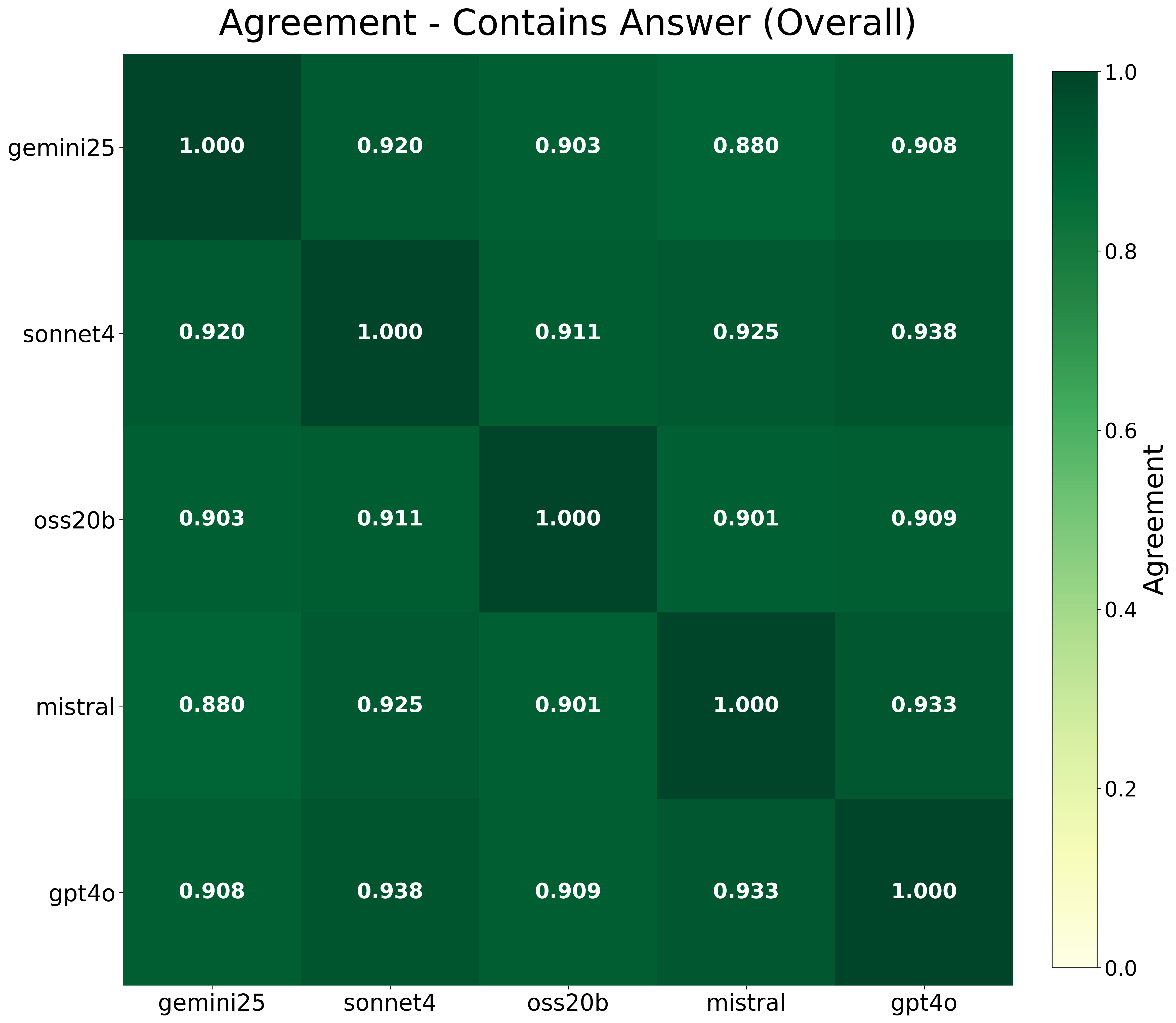}
    \caption{Pairwise agreement: contains answer}
\end{subfigure}
\caption{Inter-judge consistency across five LLM judges used to validate
the Stage~1 reward signal. Judges show strong agreement on relevance
ranking and on whether a candidate passage contains the answer, which
supports the use of judge-derived outputs as soft supervision for
teacher training.}
\label{fig:judge_agreement}
\end{figure*}

Figure~\ref{fig:judge_agreement} shows that inter-judge consistency is
strong on two core dimensions. For relevance, judges exhibit high
Spearman rank correlation, indicating that they largely preserve the
same ordering of candidate passages. For answer containment, pairwise
agreement is also high, suggesting that judges are broadly consistent in
identifying whether a passage contains the required answer signal. These
results support our use of judge-derived outputs as \emph{soft} reward
supervision for teacher training, while still treating them as an
approximation rather than a substitute for human annotation.

\begin{figure}[t]
\centering
\includegraphics[width=\columnwidth]{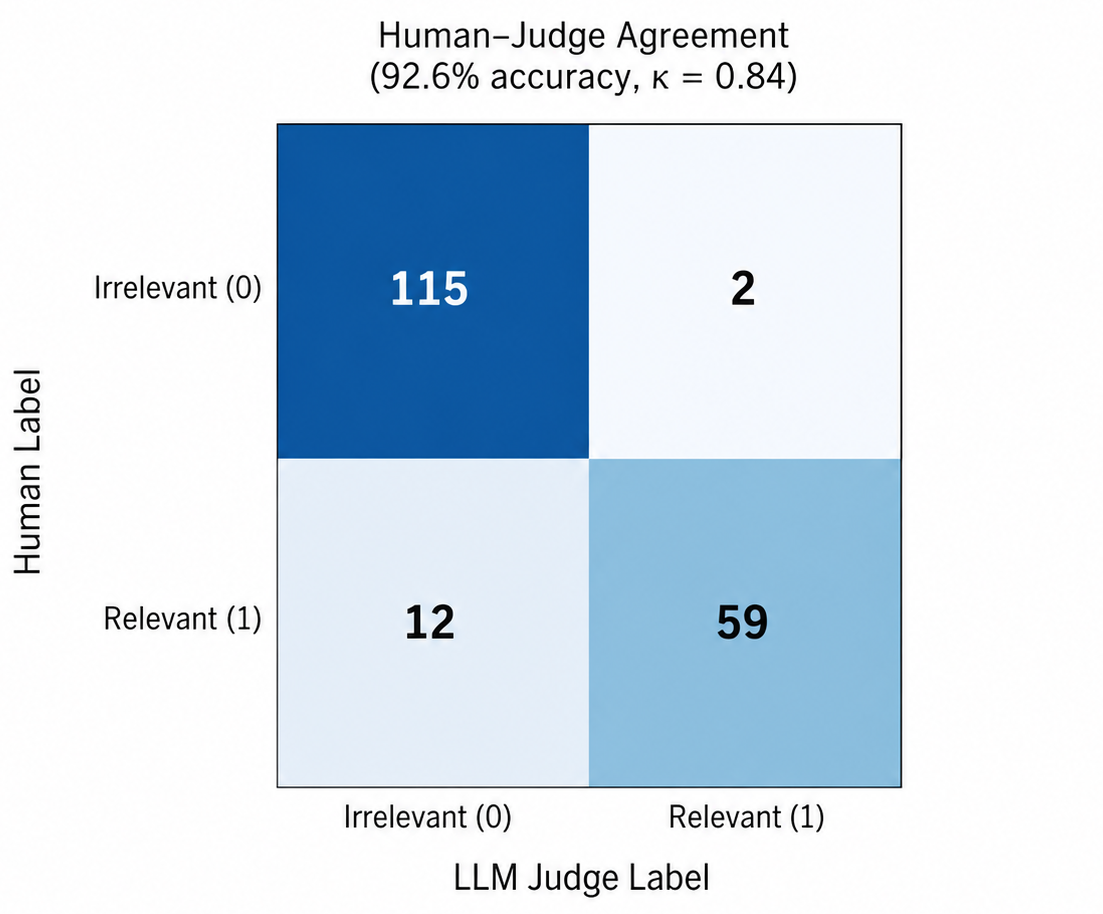}
\caption{Agreement between the Stage~1 LLM judge and human relevance
annotations. The judge achieves 92.6\% agreement and Cohen's
$\kappa=0.84$, with most disagreements coming from false negatives,
indicating a conservative bias relative to human labeling.}
\label{fig:human_judge_agreement}
\end{figure}

Beyond inter-judge consistency, we also validate the Stage~1 judge
against human annotations on a manually labeled relevance set
(Figure~\ref{fig:human_judge_agreement}). The LLM
judge achieves 92.6\% agreement with human labels (174/188 examples),
with Cohen's $\kappa=0.84$, indicating near-perfect agreement. This
result suggests that the judge is not merely self-consistent across
strong LLM evaluators, but also closely aligned with human relevance
judgments. The disagreement pattern is asymmetric: most errors are false
negatives, indicating that the judge is somewhat stricter than human
annotators rather than overly permissive.

\subsection{Per-Dataset Human--Judge Agreement}
\label{sec:appendix_human_judge}

\begin{figure*}[t]
\centering
\includegraphics[width=0.9\textwidth]{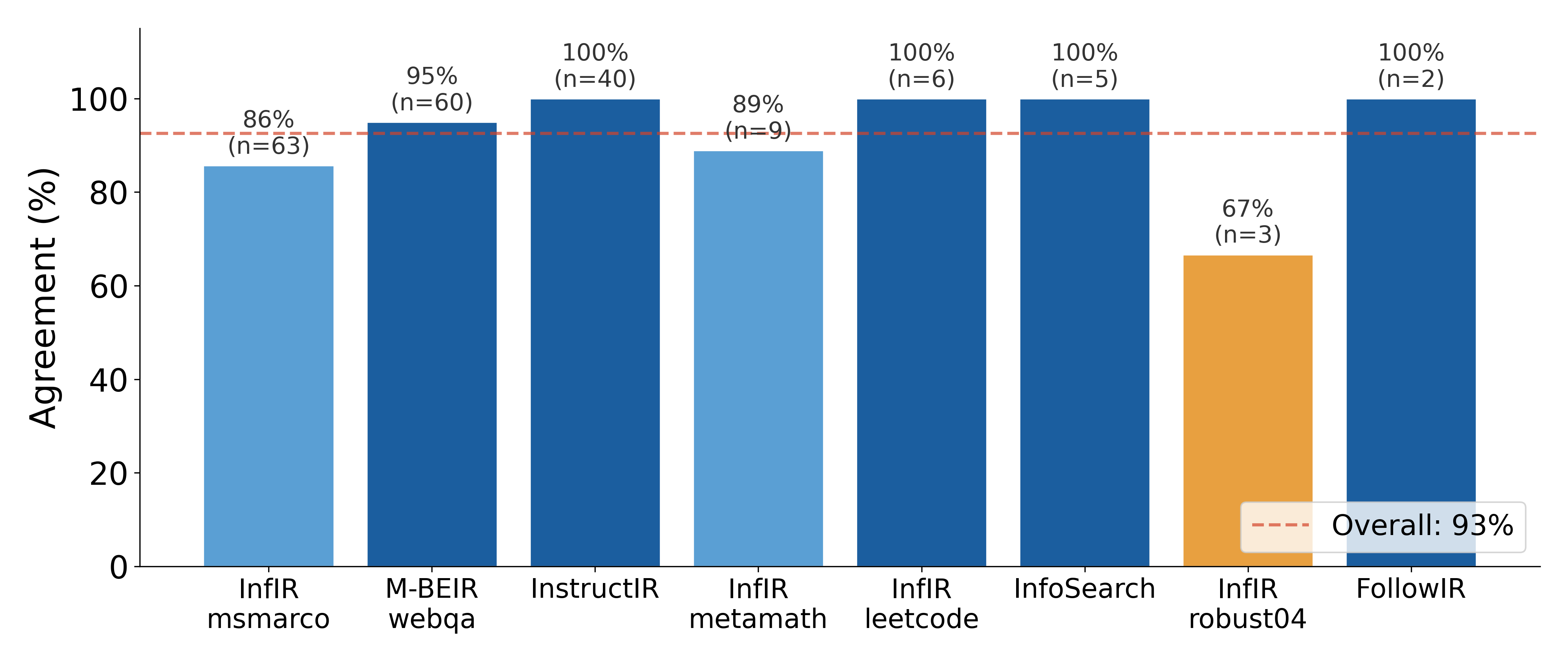}
\caption{Per-dataset agreement between the LLM judge and human
annotations. The dashed line shows the overall agreement rate of 92.6\%.
Agreement is high across most datasets, with some variation in smaller
buckets.}
\label{fig:human_judge_per_dataset}
\end{figure*}

To better understand whether judge--human agreement is concentrated in a
small subset of the evaluation data, we report agreement broken down by
dataset in Figure~\ref{fig:human_judge_per_dataset}. Agreement remains
high across most datasets, with especially strong alignment on
InstructIR, InfIR LeetCode, InfoSearch, and FollowIR. The lowest
agreement appears on InfIR Robust04, but this bucket is also small and
does not change the overall conclusion that the judge is broadly aligned
with human relevance judgments.

\section{Expanded Validation Visualizations Across All Baselines}
\label{sec:appendix_visuals}

This section provides subset-level comparisons across all major
baselines: Qwen3-4B, Base-1B, BGE v2 M3, Teacher GRPO, Cohere v3.5,
Cohere v4-fast, Jina v2, Rank-R1-7B, REARANK-7B, and A3
Distilled-1B. These plots complement the aggregate tables by showing
where models differ across datasets.

\begin{figure*}[!t]
\centering
\includegraphics[width=0.98\textwidth]{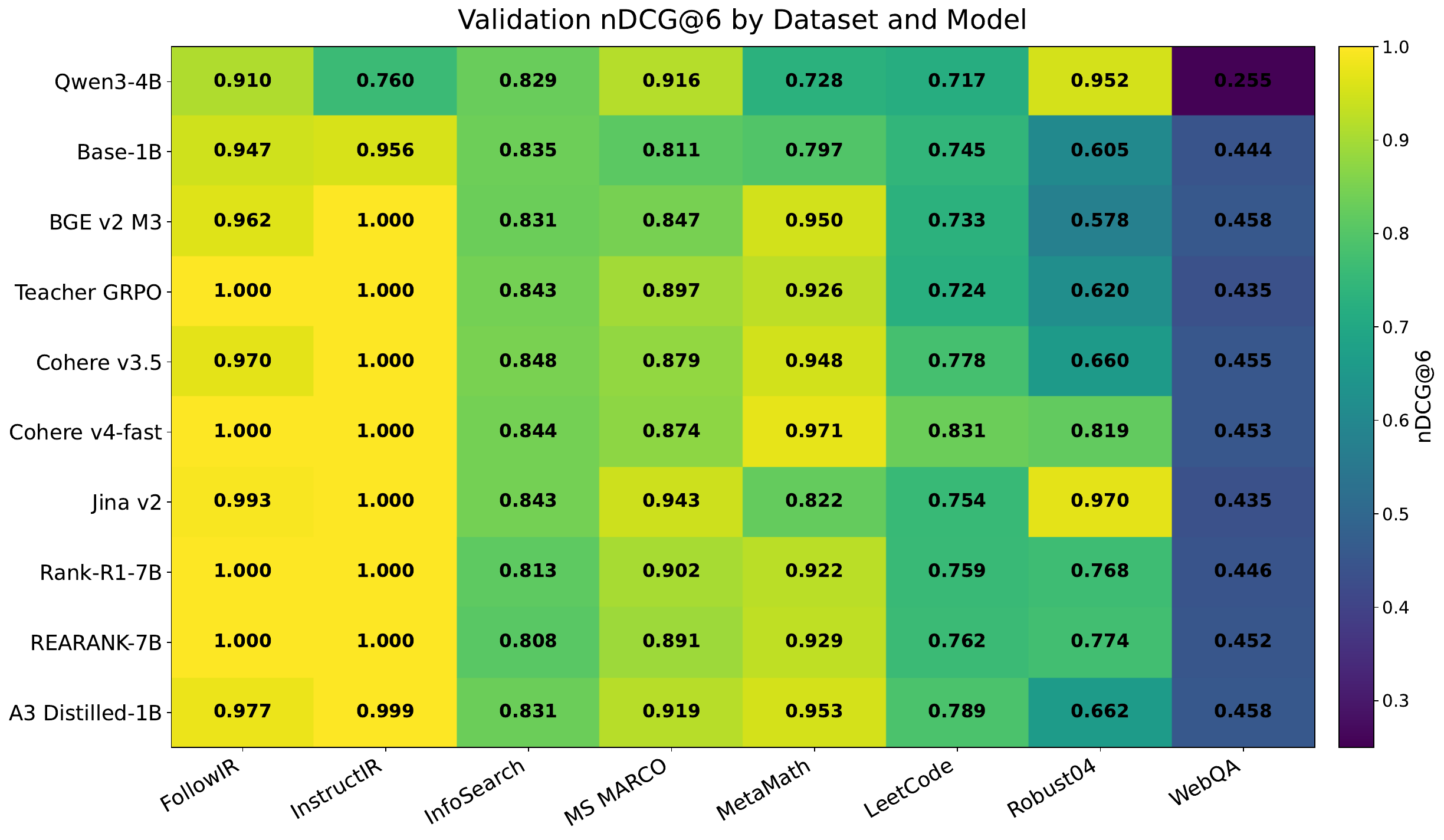}
\caption{Validation nDCG@6 by dataset and model across all major
baselines. The full heatmap makes clear that the distilled student is
competitive across a diverse range of subset types, while different
baselines exhibit distinct strengths on specific domains.}
\label{fig:appendix_full_heatmap}
\end{figure*}

\begin{figure*}[!t]
\centering
\includegraphics[width=0.92\textwidth]{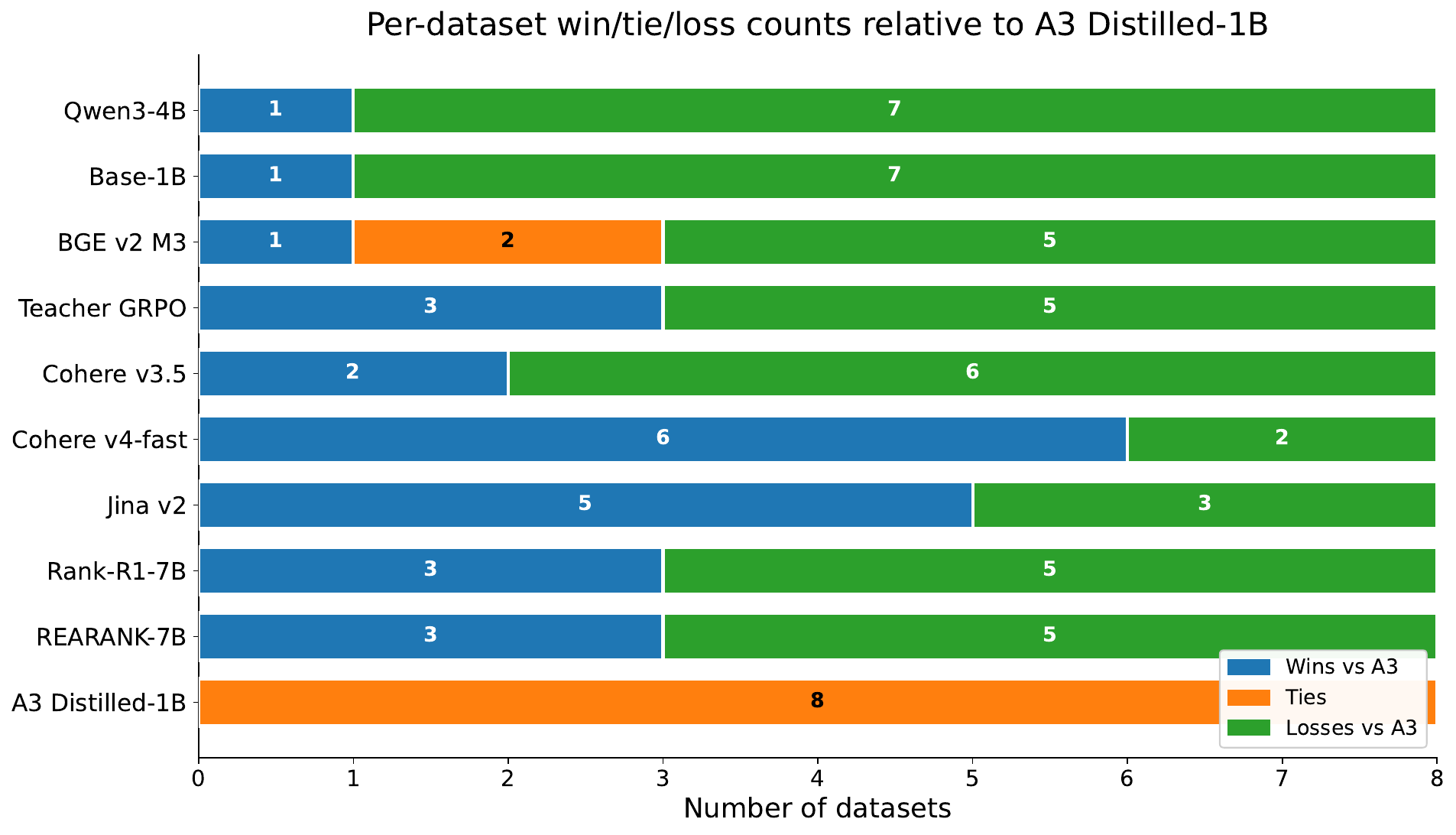}
\caption{Per-dataset win/tie/loss counts relative to A3 Distilled-1B on
the validation benchmark. The distilled student has one of the strongest
overall profiles, although different baselines still win on specific
subsets.}
\label{fig:appendix_full_wtl}
\end{figure*}

\begin{figure*}[!t]
\centering
\includegraphics[width=0.80\textwidth]{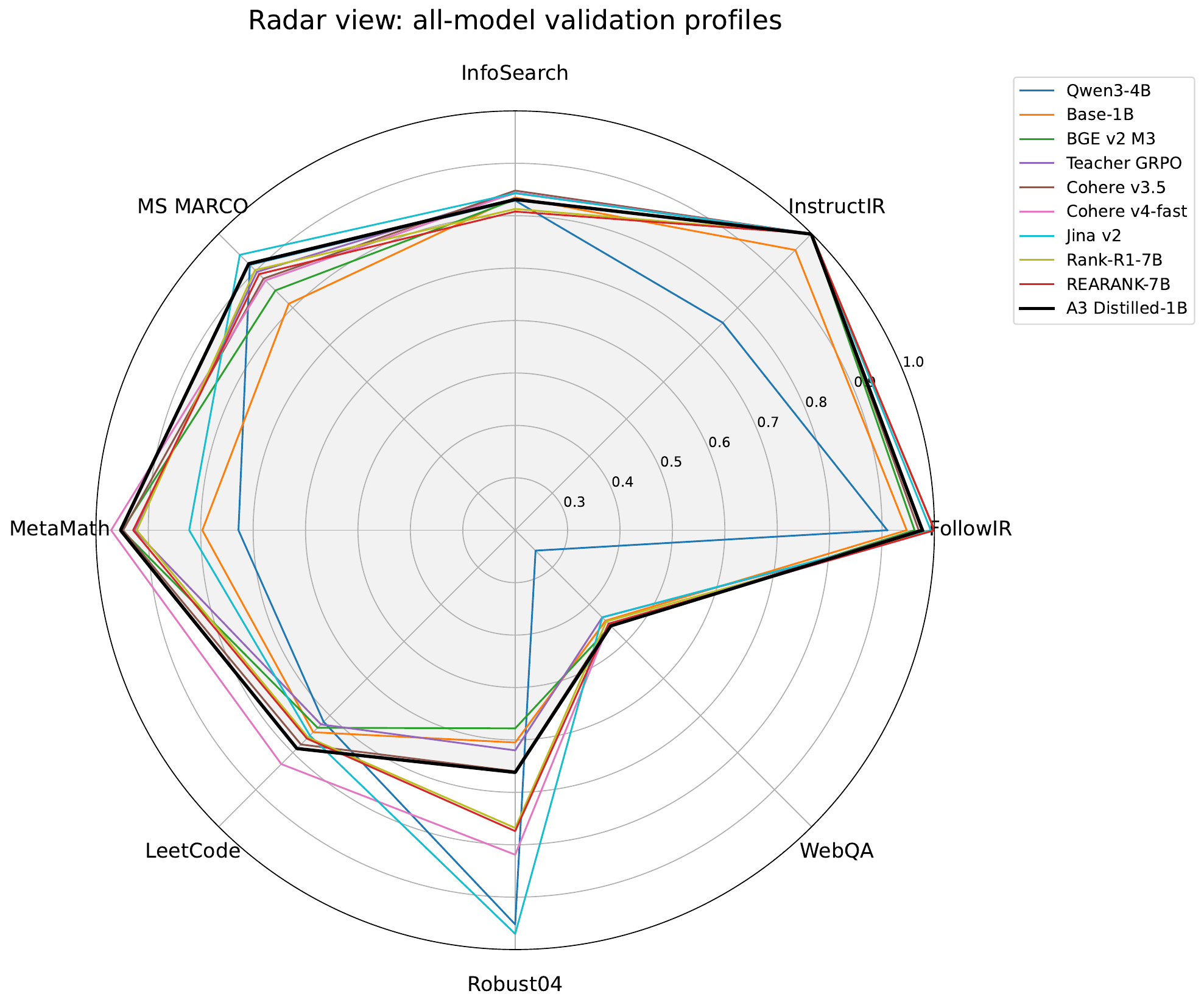}
\caption{Radar view of validation performance profiles across all major
baselines. This visualization emphasizes the overall shape of each
model's strengths and weaknesses across the 8 validation subsets.}
\label{fig:appendix_full_radar}
\end{figure*}

\begin{figure*}[!t]
\centering
\includegraphics[width=0.8\textwidth]{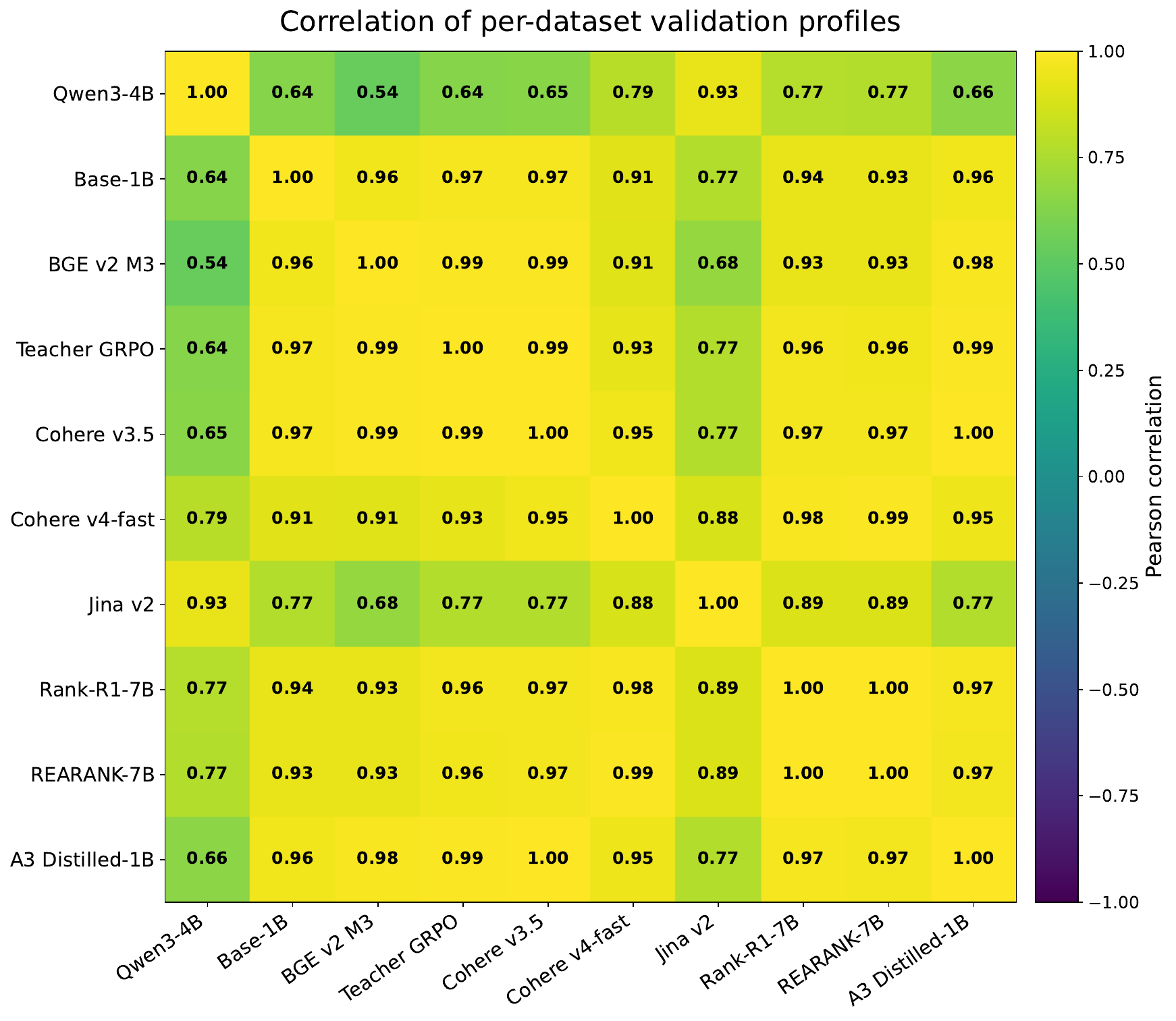}
\caption{Correlation of per-dataset validation profiles across all major
baselines. Higher correlations indicate models that behave similarly
across subsets, even when their aggregate scores differ.}
\label{fig:appendix_full_corr}
\end{figure*}

Figure~\ref{fig:appendix_full_heatmap} summarizes per-subset
performance across all models. Figure~\ref{fig:appendix_full_wtl}
reports win/tie/loss counts against A3 Distilled-1B across the 8
validation subsets. Figure~\ref{fig:appendix_full_radar} shows the
overall shape of each model's subset-level profile, and
Figure~\ref{fig:appendix_full_corr} shows correlations among those
profiles.

\end{document}